\documentclass[runningheads]{llncs}
\usepackage[T1]{fontenc}
\usepackage{adjustbox}
\newcommand{\ms}[2]{$#1_{\pm#2}$}
\usepackage{hyperref}
\usepackage[numbers,sort&compress]{natbib}\usepackage{newtxtext}    
\usepackage[utf8]{inputenc}
\usepackage{pifont}
\usepackage{booktabs}

\usepackage{amsmath}
\usepackage{tcolorbox}
\newtcolorbox{mybox}[1]{
  colback=cyan!5!white,
  colframe=cyan!75!black,
  fonttitle=\bfseries,
  title=#1,
  left=3pt,
  right=3pt,
  top=2pt,
  bottom=2pt
}
\usepackage{microtype}
\usepackage{graphicx}

\usepackage{array}
\usepackage{xcolor}

\newif\ifshowstale \showstalefalse

\usepackage{colortbl}
\usepackage{siunitx}
\usepackage{pdflscape}

\definecolor{headerblue}{RGB}{30,70,130}
\definecolor{rowgray}{RGB}{245,247,250}

\usepackage{subcaption}%
\usepackage[varvw]{newtxmath}       

\usepackage{graphicx}
\begin{document}
\title{Beyond Questions:\texorpdfstring{\\}{ }Evaluating What LLMs (Actually) Know}
%
%
\author{Luca Giordano\inst{1} \and
Simon Razniewski\inst{1}}
\authorrunning{L. Giordano et al.}
%
\institute{ScaDS.AI Dresden/Leipzig \& TU Dresden, Germany
\email{\{luca.giordano,simon.razniewski\}@tu-dresden.de}}
\maketitle              
\begin{abstract}
Parametric knowledge in large language models (LLMs) is a cornerstone of their success, yet remains poorly understood. Existing knowledge benchmarks typically rely on predefined questions (e.g., “What is the birth date of M.L. King?”), evaluating only knowledge that benchmark designers explicitly choose to query, a problematic availability bias.

In this paper, we introduce \textbf{open knowledge evaluation}, a new paradigm for benchmarking the factual knowledge of LLMs. Instead of asking narrow questions, it evaluates models on the knowledge they surface in response to open-ended elicitation prompts (e.g., “Tell me everything you know about M.L. King”). This shifts the focus from predefined answer retrieval toward characterizing the knowledge models naturally express.

We instantiate this paradigm with \textbf{BeQu} (\textbf{Be}yond \textbf{Qu}estions), a benchmark resource of 10{,}000 entities paired with reference corpora for statement verification. Using BeQu, we evaluate a broad range of language models and analyze the effects of reasoning effort, model scale, prompt format, schema type, and knowledge domain. Data and code are available in the \href{https://github.com/Knowledge-aware-AI/BeyondQuestions}{GitHub repository} (\href{https://doi.org/10.5281/zenodo.24281405}{doi:10.5281/zenodo.24281405}), as well as on the continuously updated benchmark website at {\small\url{https://bequ.scads.ai/}}.
\end{abstract}

\noindent
{\small\textbf{Resource type:} Evaluation Benchmark}\\
{\small\textbf{License:} CC-BY 4.0 International (CC-BY-SA 4.0 for Wikipedia text)}\\
{\small\textbf{DOI:} \href{https://doi.org/10.5281/zenodo.24281405}{10.5281/zenodo.24281405}}\\
{\small\textbf{URL:} {\small\url{https://bequ.scads.ai/}}
\section{Introduction}
\medskip
\begin{quote}
\itshape
The answers one gets depend\\ on the questions one asks.\\
   \mbox{\qquad\qquad} --- \textit{attributed to Thomas Kuhn}
\end{quote}

\paragraph{Motivation and problem.}
Large language models (LLMs) internalize substantial amounts of factual knowledge during pre-training \citep{bubeck2023sparks}, enabling strong performance across a wide range of downstream tasks. At the same time, they remain difficult to analyze and evaluate: their knowledge is implicit, inconsistently expressed, and affected by hallucinations, biases, and factual inaccuracies \citep{holtzman2020curious, ji2023survey, berglund2024reversal}. In practice, prompting remains the primary mechanism for accessing this embedded knowledge.

\begin{figure}
    \centering
    \includegraphics[width=1\linewidth]{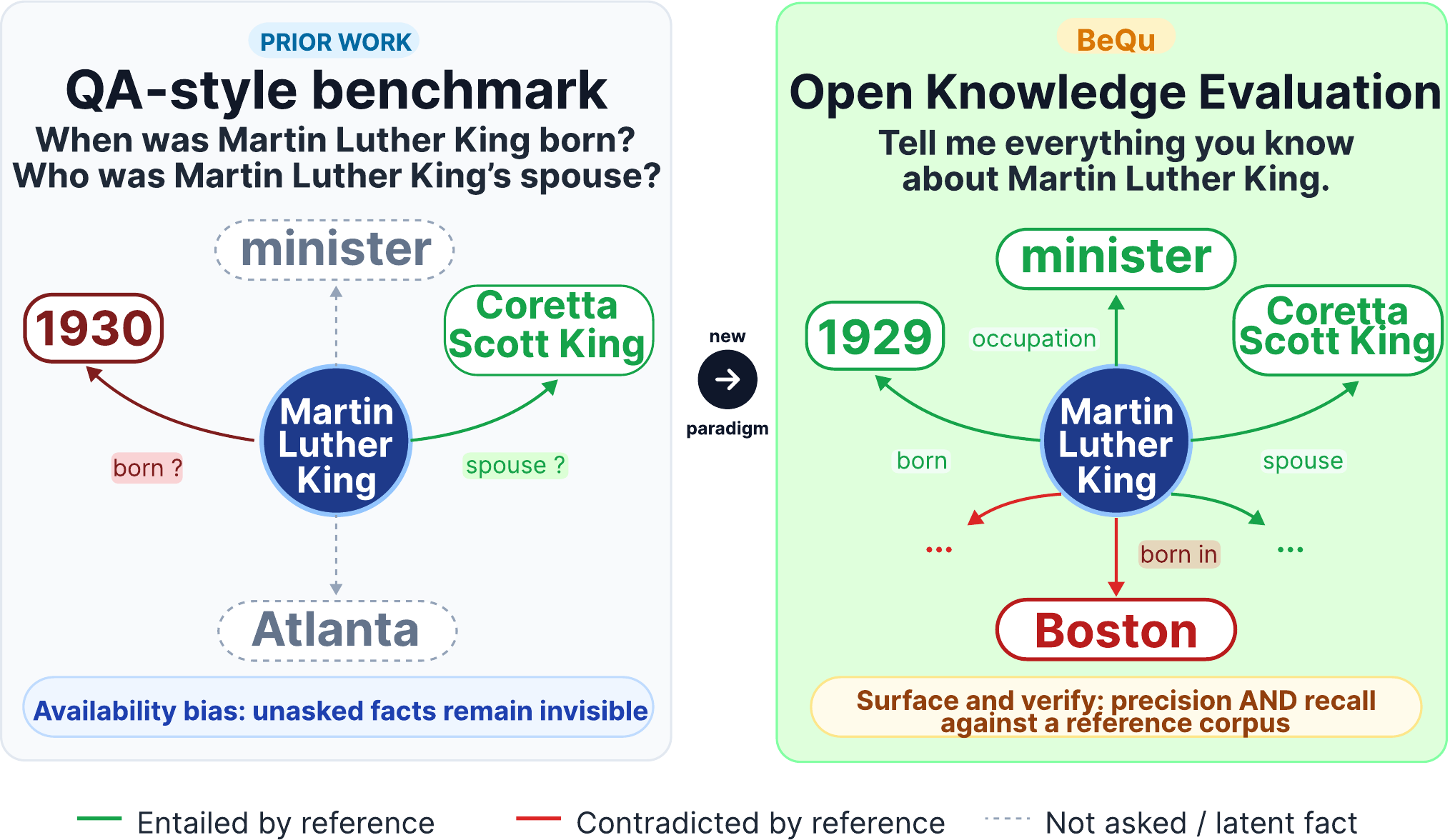}
    \caption{BeQu vs. previous benchmarks.}
    \label{fig:teaser}
\end{figure}
The dominant paradigm for evaluating LLM knowledge relies on benchmarks composed of predefined question-answer pairs. While highly influential, such benchmarks are known to suffer from important limitations, including saturation, data leakage, prompt sensitivity, and annotation biases \citep{bowman-dahl-2021-will, kiela-etal-2021-dynabench, liang2022holistic, jacovi-etal-2023-stop}. More fundamentally, they suffer from \emph{availability bias}: they can only evaluate knowledge that benchmark designers explicitly choose to query \citep{raji2021ai, bean2025measuring, hu2025enabling, giordano-razniewski-2026-foundations}. Knowledge or beliefs outside the benchmark scope remains invisible.

Recent work on \emph{knowledge materialization} addresses this limitation through recursive and open-ended extraction of model knowledge into structured knowledge bases \citep{cohen2023crawling, hu2025enabling, giordano-razniewski-2026-foundations}. Instead of asking predefined questions, these approaches elicit knowledge freely and represent it as relational statements. This line of work suggests a broader paradigm for analyzing LLM knowledge, but also raises a central open challenge: how can such open-ended knowledge elicitation itself be evaluated reliably and systematically?

\paragraph{Approach and Contribution.}
In this paper, we introduce \textbf{open knowledge evaluation}, a new paradigm for benchmarking the factual knowledge of LLMs. Instead of testing models with narrowly phrased questions (e.g., ``What is the birth date of Martin Luther King?''), open knowledge evaluation evaluates the factual statements models generate in response to open-ended prompts (e.g., ``Tell me everything you know about Martin Luther King'') (Figure \ref{fig:teaser}). This shifts the focus from predefined answer retrieval toward characterizing the knowledge models naturally surface.
We instantiate this paradigm with \textbf{BeQu} (\textbf{Be}yond \textbf{Qu}estions), a benchmark built from 10{,}000 entities paired with reference corpora for statement verification. Our main contributions are:
\begin{enumerate}
    \item We introduce \textbf{open knowledge evaluation}, a novel paradigm for benchmarking LLM knowledge through open-ended elicitation rather than predefined question-answer pairs.
    
    \item We present \textbf{BeQu} (\textbf{Be}yond \textbf{Qu}estions), a benchmark implementing this paradigm through open-ended prompts and reference-based verification of generated knowledge statements.
    
    \item Using \textbf{BeQu}, we conduct a comprehensive empirical study across a broad range of language models and settings, analyzing the effects of reasoning effort, model scale, prompt format, and knowledge domain on knowledge expression and factual accuracy.
\end{enumerate}

Our main empirical findings are as follows: Commercial models still produce the highest-quality knowledge, although several open-weight models come remarkably close. Reasoning effort impacts knowledge quality differently depending on models. Enforcing standardized output schemas substantially improves precision, but at the cost of recall. Model behaviour is stable, with low standard deviations in results across runs. Finally, prompt-level attempts to explicitly trade precision for recall appear to have only limited effect.

Data and code are available under the CC-BY 4.0 license (CC-BY-SA 4.0 for Wikipedia text) in the \href{https://github.com/Knowledge-aware-AI/BeyondQuestions}{GitHub repository}, as well as on the continuously updated benchmark website at {\small\url{https://bequ.scads.ai/}} (Figure \ref{fig:website}; for details, see Appendix - \ref{app:benchmark_website}).

\begin{figure}[ht!]
    \centering
    \includegraphics[width=1\linewidth]{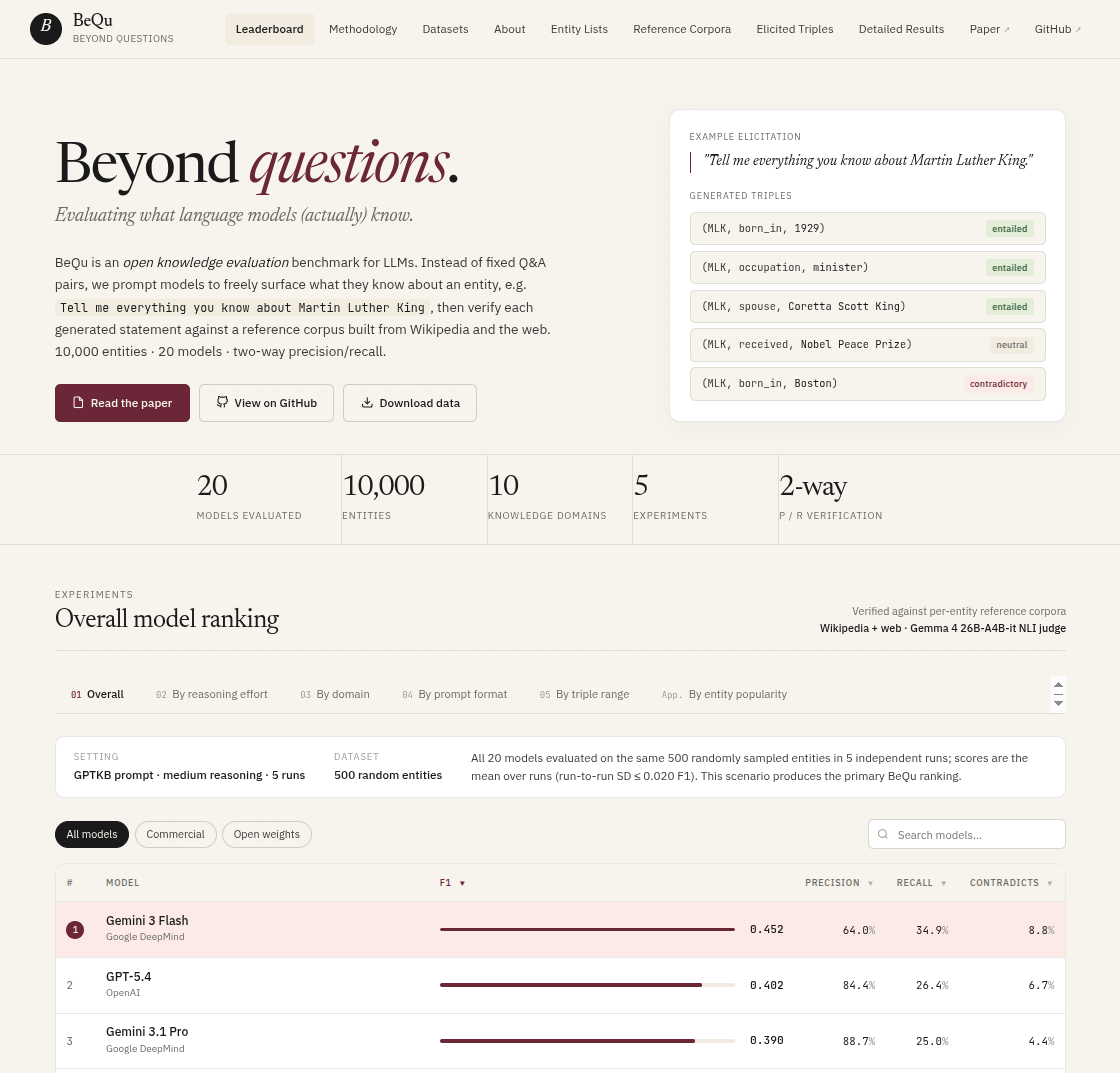}
    \caption{Website homepage of the BeyondQuestions benchmark.}
    \label{fig:website}
\end{figure}

\section{Background}

\paragraph{Q\&A Knowledge Benchmarks.}
The dominant paradigm for evaluating LLM knowledge relies on benchmarks composed of predefined question-answer pairs, differing in scope, format, and difficulty but sharing a common assumption: that knowledge can be assessed through a fixed set of queries and matching reference answers.

General-purpose benchmarks, e.g. MMLU \citep{hendrycks2020measuring}, BIG-bench \citep{srivastava2023beyond}, Natural Questions \citep{kwiatkowski-etal-2019-natural}, and Humanity's Last Exam \citep{center2026benchmark}, evaluate broad world knowledge and reasoning across diverse domains, while others focus on specialized domains, such as medicine (MedQA \citep{yao-etal-2026-medqa}), law (LegalBench \citep{guha2023legalbench}), and finance (FinanceBench \citep{islam2023financebench}).

Despite their importance, existing knowledge benchmarks suffer from several fundamental limitations \citep{bowman-dahl-2021-will, kiela-etal-2021-dynabench, liang2022holistic, jacovi-etal-2023-stop}:
\begin{enumerate}
    \item \textbf{Saturation:} Many widely used benchmarks are approaching saturation, with state-of-the-art models achieving near-ceiling performance. This makes it increasingly difficult to distinguish between models and to interpret improvements meaningfully.
    \item \textbf{Leakage:} QA-benchmarks rely on a well-defined ground truth, which, for replicability, is normally provided online. This data can be scraped for the training data of future LLMs, leading to unintentional or intentional overfitting or benchmark gaming.
    \item \textbf{Availability bias:} QA-benchmarks created manually, or with humans-in-the-loop, are biased from their creator's choices. This yields a classic availability bias \cite{kahnemann}. A model may possess substantial knowledge that is never measured simply because no corresponding question was included. Conversely, strong performance on a benchmark does not imply broad knowledge.
\end{enumerate}
In contrast, our open knowledge evaluation paradigm is easy to extend to extremely long-tail topics, does not contain narrow ground truth answers that can be easily overfitted to, and does not prescribe narrow questions.

\paragraph{Evaluating Factual Knowledge Without Predefined Queries.} Other works evaluate the factual accuracy of freely generated text rather than answers to fixed questions, a setting closer to ours. FactScore \citep{min-etal-2023-factscore} decomposes long-form model outputs into atomic claims and verifies each against a retrieval corpus, yielding a fine-grained precision estimate for open-ended generation. LongFact \citep{wei2024long} and VeriScore \cite{song2024veriscore} extend this to longer and indivisible outputs, similarly measuring factual precision of free-form generation against reference evidence. These works share our interest in evaluating what models assert freely; however, they evaluate only the precision of model outputs, and do not measure recall. BeQu builds on this line of work by introducing both precision and recall as evaluation axes, as well as by grounding evaluation in a large-scale reference corpus, and by systematically studying how several dimensions shape the generated knowledge and its accuracy.

\paragraph{Knowledge Materialization.}
Knowledge materialization, as presented in \citep{cohen2023crawling} and \citep{hu2025enabling}, and further developed and explored in \citep{giordano-razniewski-2026-foundations}, \citep{saeed2026llmpedia}, and \citep{hu2026constructing} is a prompt-based, recursive, and open-ended knowledge extraction process that stores results in a structured format, such as a knowledge base (KB).

An LLM is prompted to return knowledge about a seed entity in the form of \textit{(s, p, o)} triples. Using the model itself for Named Entity Recognition (NER), new named entities are iteratively identified among the objects of these triples. These objects become new subjects and the graph is further expanded recursively. 
The key insight is to extract LLM knowledge/beliefs without picking specific queries, thus overcoming the availability bias.

The paradigm above has been utilized at scale, e.g., creating GPTKB-v1.5 containing 100M triples for 6M entities, at an API cost of over \$10k. However, their accuracy evaluation so far has been limited to post-hoc sampling of a few triples against transient web search APIs, and fundamental design decisions such as model choice, prompt format, or precision-recall tradeoffs have not been systematically investigated. This calls for a new knowledge evaluation paradigm.

\section{Open Knowledge Evaluation Paradigm}
\label{section_paradigm}

The core idea of our proposed open knowledge evaluation is to treat the model's factual knowledge as the object of evaluation, rather than asking whether a model can answer a fixed query. This shifts the focus from narrow answer retrieval to knowledge expression: which facts a model surfaces, how accurately it states them, and how much of the reference knowledge it covers.

The paradigm consists of four stages: (i) selecting evaluation entities, (ii) constructing a reference corpus, (iii) eliciting open-ended knowledge about them, and (iv) verifying elicited statements against that corpus. It supports two complementary evaluation directions. Precision measures whether model-generated statements are supported by the reference corpus, while recall measures how much reference knowledge is recovered in the model's output.

\section{The BeQu Benchmark}
\begin{figure*}
    \centering
    \includegraphics[width=0.9\linewidth]{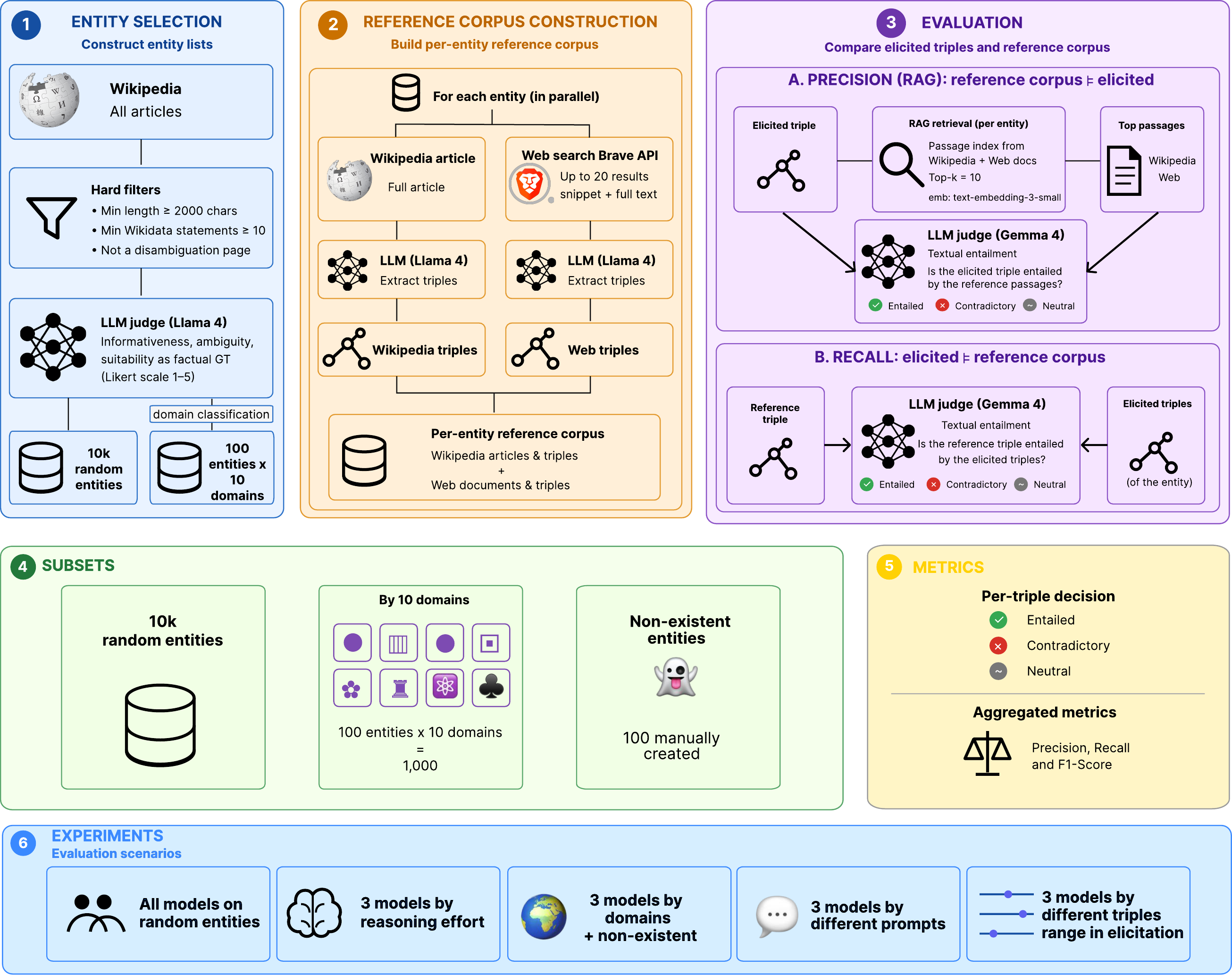}
    \caption{Overview of the BeQu benchmark.}
    \label{fig:workflow}
\end{figure*}

We instantiate the open knowledge evaluation paradigm with \textbf{BeQu} (\textbf{Be}yond \textbf{Qu}estions), a benchmark for evaluating factual knowledge expressed by LLMs. BeQu operationalizes the paradigm with 10k entities, and 6 GBs of reference corpora. Figure~\ref{fig:workflow} provides a schematic overview of our benchmark.

BeQu is constructed through entity selection and reference corpus construction. Three different subsets are defined, to shed further light on LLM knowledge.

\subsection{Entity Selection}
The first key aspect of BeQu is the set of entities we want to use to extract knowledge from the LLMs tested (part 1 in Figure \ref{fig:workflow}). We define three subsets: (i) random, (ii) domains, (iii) non-existent.

The most important one is the \textbf{random subset}, constructed by randomly sampling 10,000 Wikipedia article titles. We apply a set of hard filters: each article must have a minimum length of 2,000 characters, at least 10 statements on its respective Wikidata page, and must not be a disambiguation page. Furthermore, we filter each candidate with an LLM judge (Llama 4 Scout). The judge is prompted to assess three dimensions on a Likert scale of 1–5: informativeness (how much factual content the entity offers), ambiguity (whether the entity name is unambiguous), and suitability as a factual ground truth for LLM evaluation (prompt used in Appendix - Figure \ref{fig:judge_prompt_entities}). Any entity with any score below 3 is excluded. We validate this step against human ratings (Appendix - \ref{app:error_analysis_entity_selection}).

Both filter layers are applied until 10,000 suitable entities are collected. This first subset represents the primary benchmark subset, and is used for most of the experiments described in Section \ref{section_experiments}.

We construct the \textbf{domain subset} with the same methodology, but the LLM judge is additionally instructed to retain the candidate entity only if it belongs to (only) one of the following ten knowledge domains: \textit{person, organisation, location, event, work of art, artifact, scientific concept, cultural concept, animal}, and \textit{plant}. The random sampling continues until 100 entities per domain are collected.

Finally, we construct the \textbf{nonexistent subset} by manually defining 100 non-existent entities. This entity list includes both plausible and absurd entities e.g. \textit{Valdora Strait, International Airport of Andorra, U-Bahn Dresden, National Archive of Future Events, iPhone 19 Pro, Mediterranean Logistics Coordination Agency, Helios Prize for Digital Arts, Milan Flying Public Transport System, Gulf of Varennes, Zittau International Agreement on AI Governance}.

\subsection{Reference Corpus Construction}
\label{corpus_construction}
For each entity in subsets 1 and 2, we build a corpus that serves as the reference to verify elicited triples (part 2 in Figure \ref{fig:workflow}). Each corpus is assembled from multiple complementary sources processed in parallel.

The first source is the full \textbf{Wikipedia article} for the entity. The other sources consist of \textbf{up to 20 web documents} retrieved via the Brave Search API\footnote{\href{https://brave.com/search/api/}{https://brave.com/search/api/}}; for each web result, we collect both a short snippet and the full page text. Using GPT-OSS-120B as judge, we establish that a 20-document corpus contains, averaged across 100 sampled entities, 55\% of the information present in a 50-document corpus, and 41\% of that present in a 200-document corpus (maximum allowed by the API), thus representing a computationally efficient and representative sample (details in Appendix - \ref{app:error_analysis_coverage_staleness}).

Both the Wikipedia article and each web snippet are then processed by an LLM (Llama 4 Scout), which extracts factual triples from the text, a task where LLMs showed promising performance already back in 2023 \citep{wadhwa-etal-2023-revisiting}.

\paragraph{Resource statistics.} Both resources (random and by domain) were collected in March 2026. The resource for the 10k random entities (subset 1) comprises 152,808 documents and 1.10M reference triples; All entities are linked to their Wikidata ID. On average, each entity is covered by its Wikipedia article (623 words; median 311) and 15.3 web documents (median 16; 5th–95th percentile 10–19), totalling about 19k words of web text per entity and 197M words overall. The web documents come from 36,346 distinct domains: Wikipedia and Wikidata pages make up 21\% of them, and no other domain exceeds 3\%. From these sources, Llama 4 Scout extracts on average 110 reference triples per entity (median 108): 16.8 from the Wikipedia article and 93.7 from the web search snippets. The 20 most frequent predicates (e.g., \texttt{type}, \texttt{locatedIn}, \texttt{occupation}, \texttt{birthDate}) cover 15\% of the triples. For precision, the texts are split into 4.64M passages of at most 500 characters (464 per entity on average). Alongside these resources, we release the 1.98M triples elicited in Experiments 1-5 (Section \ref{section_experiments}) and Additional Experiment 2 (Appendix \ref{app:additional_exp_2}), and 387,732 judged triples, each with the judge's label and explanation.

The resource for the entities by domain (subset 2) has a similar profile: 15,156 web documents (15.2 per entity) from 5,513 distinct domains (Wikipedia and Wikidata: 19\%), about 20M words, and 91,872 reference triples (92 per entity: 14.7 from Wikipedia and 77.2 from the web), split into 449k passages (449 per entity); all 1,000 entities are linked to their Wikidata ID.

\subsection{Evaluation}
\label{section_evaluation}
BeQu evaluates models along two complementary axes: \textbf{precision}, which measures how well model-elicited triples are supported by the reference corpus, and \textbf{recall}, which measures how much of the reference corpus knowledge the model surfaces (part 3 in Figure \ref{fig:workflow}). The evaluation is configured as a textual entailment a.k.a. natural language inference (NLI) task for both directions, with a separate, locally-hosted Gemma 4 26B-A4B-it serving as the NLI judge. We tested 7 different models (GLM 5.3 Flash, Gemma 4 26B A4B, Llama 3.3 70B, Qwen 3.8 27B, MiniMax M3, DeepSeek V4.1 Flash and GPT-OSS-120B) and chose Gemma 4 26B, because in our experiments it both has among the highest agreement with human (80\% accuracy in 3-level rating entail/neutral/contradict), good agreement with others (average pairwise kappa vs. all 6 others is 0.721), and is comparably cheap/small in terms of parameters.\footnote{Our task is similar to FEVER \citep{fever-2026-1}, where LLMs reached $>$92\% accuracy in 2024. This is also why the literature mentions no numbers for newer models (like Gemma 4 26B-A4B-it) on this task anymore. We conducted a small human evaluation on 100 triples for precision and 100 for recall, finding 80\% agreement (84\% precision, 75\% recall) with humans (who are themselves imperfect).}

\textbf{Precision (elicited $\rightarrow$ reference corpus)}. For each elicited triple, we retrieve the top-10 most relevant passages from the per-entity passage index using a RAG pipeline. Passages are created by splitting texts at double newlines (\texttt{\textbackslash n\textbackslash n}) and splitting longer paragraphs into chunks of at most 500 characters at word boundaries, so that the whole reference corpus can be retrieved. Retrieval uses the \textit{text-embedding-3-small} embedding model by OpenAI\footnote{We also test a local instance of \textit{all-mpnet-base-v2}, but qualitatively observe worse matching retrieved passages and a slightly lower rate of textual entailment. However, the BeQu benchmark still allows to easily plug-in any open-weights embedding model.}. The retrieved passages and the elicited triple are then passed to the Gemma 4 26B-A4B-it judge, which classifies the triple as Entailed, Contradictory, or Neutral with respect to the passages. We instruct the judge to classify as Neutral if the passages contain conflicting information (prompt in blue box below).

We further validate the precision evaluation by submitting to the judge the Wikipedia reference triples themselves, reaching 91.5\% precision. As a negative control, Wikipedia triples whose object is replaced by the object of a triple with the same predicate of another entity reach only 2.2\% precision. Details in Appendix \ref{app:validation_precision_judge}.

\textbf{Recall (reference corpus $\rightarrow$ elicited)}. For each reference triple, the LLM judge is asked whether the triple is entailed by the full set of elicited triples for that entity. The judge again returns one of the three entailment labels (prompt in blue box below).

Aggregate metrics are computed as entailment precision (fraction of elicited triples labelled Entailed), entailment recall (fraction of reference triples labelled Entailed), and entailment F1-Score (harmonic mean of precision and recall). For the experiment on the third (nonexistent) subset, we just count how many triples are generated (there should be zero). An example of open knowledge evaluation is reported in Figure \ref{fig:example} (Appendix - \ref{app:eval_example}).

\begin{mybox}{LLM Judge Prompt -- Precision}
    ``Can the given RDF triple be inferred from the given retrieved passages?\\
    a) The passages entail the RDF triple.\\
    b) The passages contradict the RDF triple.\\
    c) The truth of the given RDF triple cannot be determined from the passages alone.\\
    If the provided evidence is internally CONFLICTING (i.e., it contains information that both supports AND contradicts the statement), you MUST answer `c' (neutral), rather than picking a side. Respond with JSON containing `answer' (one of a/b/c) and `reasoning'.''\\[4pt]
    ``Statement to verify: \textit{(subject, predicate, object)}.''\\[4pt]
    ``Retrieved ground truth passages: \textit{[top-10 passages of at most 500 characters, each prefixed with its source, \texttt{[WIKIPEDIA]} or \texttt{[WEB]}]}''
\end{mybox}

\begin{mybox}{LLM Judge Prompt -- Recall}
    ``Can the given RDF triple be inferred from the given list of triples?\\
    a) The list of triples entails the given RDF triple.\\
    b) The list of triples contradicts the given RDF triple.\\
    c) The truth of the given RDF triple cannot be determined from the list of triples alone.\\
    If the provided evidence is internally CONFLICTING (i.e., it contains information that both supports AND contradicts the statement), you MUST answer `c' (neutral), rather than picking a side. Respond with JSON containing `answer' (one of a/b/c) and `reasoning'.''\\[4pt]
    ``Statement to verify: \textit{(subject, predicate, object) of the reference triple}.''\\[4pt]
    ``List of triples to verify from: \textit{[all triples the model elicited for the entity]}''
\end{mybox}

\section{Experiments}
\label{section_experiments}

We structure our empirical evaluation around five experimental scenarios, each designed to isolate a specific factor that may influence how much accurate knowledge a model surfaces: (1) \textbf{All models on random entities.} All models are evaluated on the first subset, with 5 runs to estimate run-to-run variability. This scenario produces the primary model ranking. (2) \textbf{By reasoning effort.} Three reasoning models are evaluated on the first subset across different reasoning effort settings (low, medium, high). (3) \textbf{By domain + non-existent.} Three models are evaluated on the second subset and, additionally, we prompt them to generate knowledge on the third subset. (4) \textbf{By prompt format.} Three models are evaluated under six prompt variations on the first subset. (5) \textbf{By triple range.} Three models are evaluated on the first subset with increasing numbers of requested triples per entity in the GPTKB prompt. 

\paragraph{Prompt format.} The prompt used in all experiments is always the prompt from the GPTKB experiments conducted in \citep{hu2025enabling} (blue box below). In Experiment 5, we modulate the range of triples requested per entity, as we hypothesize it affects the precision-recall tradeoff. However, in Experiment 4 we also test 6 prompt format variations: \textbf{1.} \textit{GPTKB} \textbf{2.} \textit{LMCRAWL} \textbf{3.} \textit{Wikidata Schema} \textbf{4.} \textit{Wikidata Schema (no constraints)} \textbf{5.} \textit{Schema.org Schema} \textbf{6.} \textit{Schema.org Schema (no constraints)}. 
Unlike GPTKB, which asks for all facts about the subject in a single prompt, LMCRAWL \citep{cohen2023crawling} elicits knowledge in two steps: it first asks the model to list the relations relevant to the subject, then queries the objects for each relation separately, using few-shot examples and allowing the model to answer “don’t know.”
In the Wikidata and Schema.org variations, we modify the GPTKB prompt to adhere to the two respective schemas (whereas the normal GPTKB prompt does not enforce any schema). For both variations we then create another version where we instruct the model not only to adhere to the schema specified, but also to a list of predefined predicates we provide in the prompt.\footnote{We generate the lists of predefined predicates by prompting GPT-5.4 for the core semantically relevant and most popular properties of the two schemas respectively. We are not highly concerned about the quality or coverage of the predicate lists themselves, but rather whether the tested models will actually follow the instructions given and whether this will affect the precision-recall tradeoff, as we hypothesize.}

\begin{mybox}{GPTKB Prompt \citep{hu2025enabling}}
    "You are a knowledge base construction expert. Given a subject entity, return all facts that you know for the subject as a list of (subject, predicate, object) triples. The number of facts may be very high, between 50 to 100 or more, for very popular subjects. For less popular subjects, the number of facts can be very low, like 5 or 10. Important:\\
    \begin{itemize}
        \item If you don't know the subject, return an empty list. 
        \item If the subject is not a named entity, return an empty list.
        \item If the subject is a named entity, include at least one triple where predicate is "instanceOf".
        \item Do not get too wordy. 
        \item Separate several objects into multiple triples with one object."
    \end{itemize}
\end{mybox}

\paragraph{Model selection.} Between September and October 2026 we evaluate a total of \textbf{20 models} spanning all major model families, several providers, parameter sizes, both open-source and commercial models.\footnote{In Experiments 1 and 3, all models which expose a reasoning effort parameter are set to Medium; Experiment 2 varies it explicitly. In Experiments 4 and 5 the reasoning parameter is left at the API default, which is no reasoning for the evaluated models. Since Experiments 4 and 5 compare prompt variants within the same configuration, their relative comparisons are unaffected, but their absolute scores are not directly comparable to those of Experiment 1.} For details about model access and cost and time spent, refer to Appendix \ref{app:cost} and \ref{app:time}.

From Anthropic, we evaluate \textbf{Claude Opus 4.6, Claude Sonnet 4.6}, and \textbf{Claude Haiku 4.5}. From OpenAI: \textbf{GPT-5.4, GPT-5 Mini, GPT-5 Nano}, and \textbf{GPT-OSS-120B}. From Google: \textbf{Gemma 3 4B, Gemma 3 12B, Gemma 3 27B, Gemini 3 Flash, Gemini 3.1 Flash Lite}, and \textbf{Gemini 3.1 Pro}. Additional models include \textbf{DeepSeek V3.2, MiniMax M2.5, Mistral Large 3, Kimi K2.5, Qwen 3.5 27B, Grok 4.3}, and \textbf{Llama 4 Scout 17B 16E}.

\paragraph{Computational Constraints.}
To keep the experiments tractable, we sample a set of 500 random entities from the first subset\footnote{Sample available at  \href{https://github.com/Knowledge-aware-AI/BeyondQuestions/blob/main/ENTITY_LISTS/RANDOM/EXPERIMENTS_500_random_entities_wikipedia.json}{this link}.}.
We also sample 1000 triples per model per direction using entity-then-triple stratified sampling with a fixed seed: we repeatedly draw an entity uniformly at random and then sample one triple uniformly from that entity's pool. This ensures that all entities contribute equally to the aggregate metrics, regardless of how many triples a model produces per entity.

\section{Results}
\subsection{Experiment 1: Overall Model Ranking}
\paragraph{Model ranking by F1-Score.}
\begin{figure}
    \centering
    \includegraphics[width=0.85\linewidth]{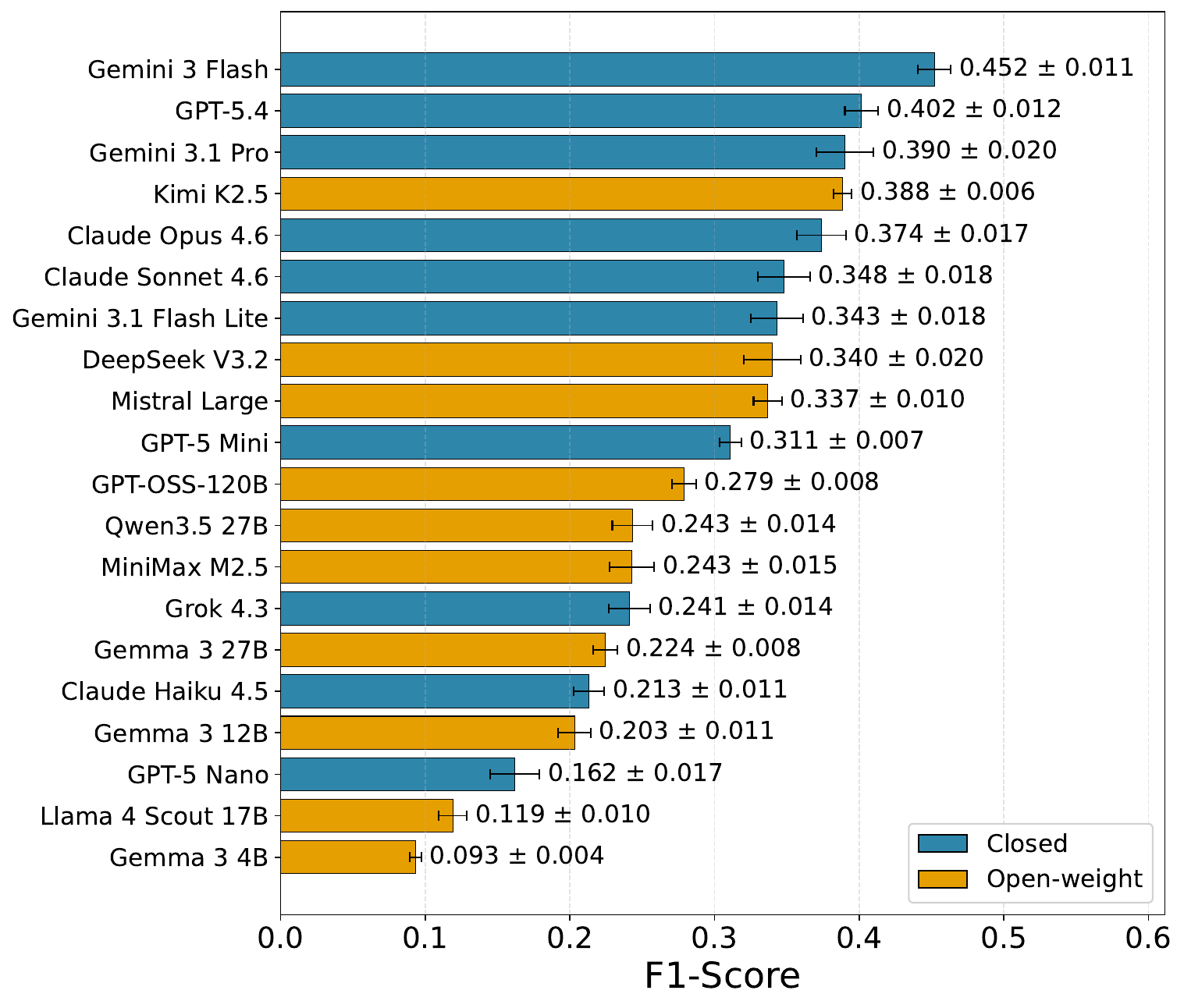}
    \caption{Experiment 1 - Model ranking by F1-Score.}
    \label{fig:exp1_1}
\end{figure}
Full results can be found in Table \ref{app:full_results_exp1} (Appendix - \ref{app:full_results}). F1-Scores are reported in Figure \ref{fig:exp1_1}. Overall mean F1 across the 5 runs spans a wide range from 0.093 (Gemma 3 4B) to 0.452 (Gemini 3 Flash). Gemini 3 Flash is the best model (F1 = 0.452), followed by GPT-5.4 (F1 = 0.402) and Gemini 3.1 Pro (F1 = 0.390), while the bottom three are GPT-5 Nano (F1 = 0.162), Llama 4 Scout (F1 = 0.119), Gemma 3 4B (F1 = 0.093). Neighbouring models are often not separable: testing each model against the next (Welch's t-test over the 5 runs, $p<0.05$) yields nine tiers, with Gemini 3 Flash alone in the top tier, followed by a second tier (GPT-5.4, Gemini 3.1 Pro, Kimi K2.5, Claude Opus 4.6, F1 0.37--0.40) and a third tier (Claude Sonnet 4.6 to Mistral Large, F1 0.33--0.34).
Three open-weight models rank in the top ten: Kimi K2.5 4th, DeepSeek V3.2 8th, and Mistral Large 9th. Commercial models still hold the top three places; the Claude variants rank 5th, 6th, and 16th.

Output verbosity does not drive the ranking: triples per entity and F1 are uncorrelated (Pearson r = 0.06, p = 0.79; Appendix \ref{app:verbosity}).

\paragraph{Precision-Recall tradeoff: the frontier boundary.}
\begin{figure}
    \centering
    \includegraphics[width=1\linewidth]{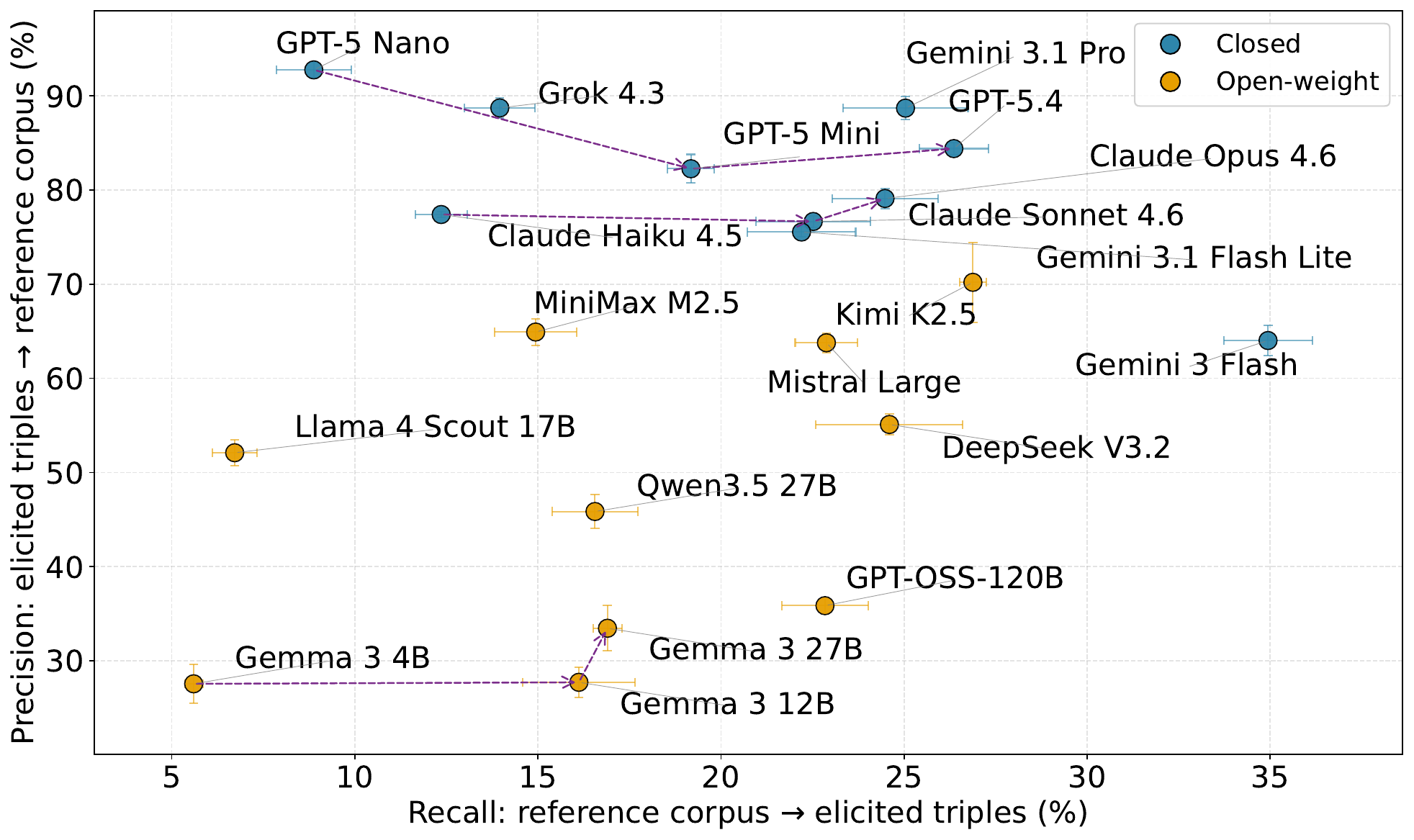}
    \caption{Experiment 1 - Precision-recall tradeoff. Dotted arrows connect models within model family by growing size.}
    \label{fig:exp1_2}
\end{figure}
\begin{figure}
    \centering
    \includegraphics[width=1\linewidth]{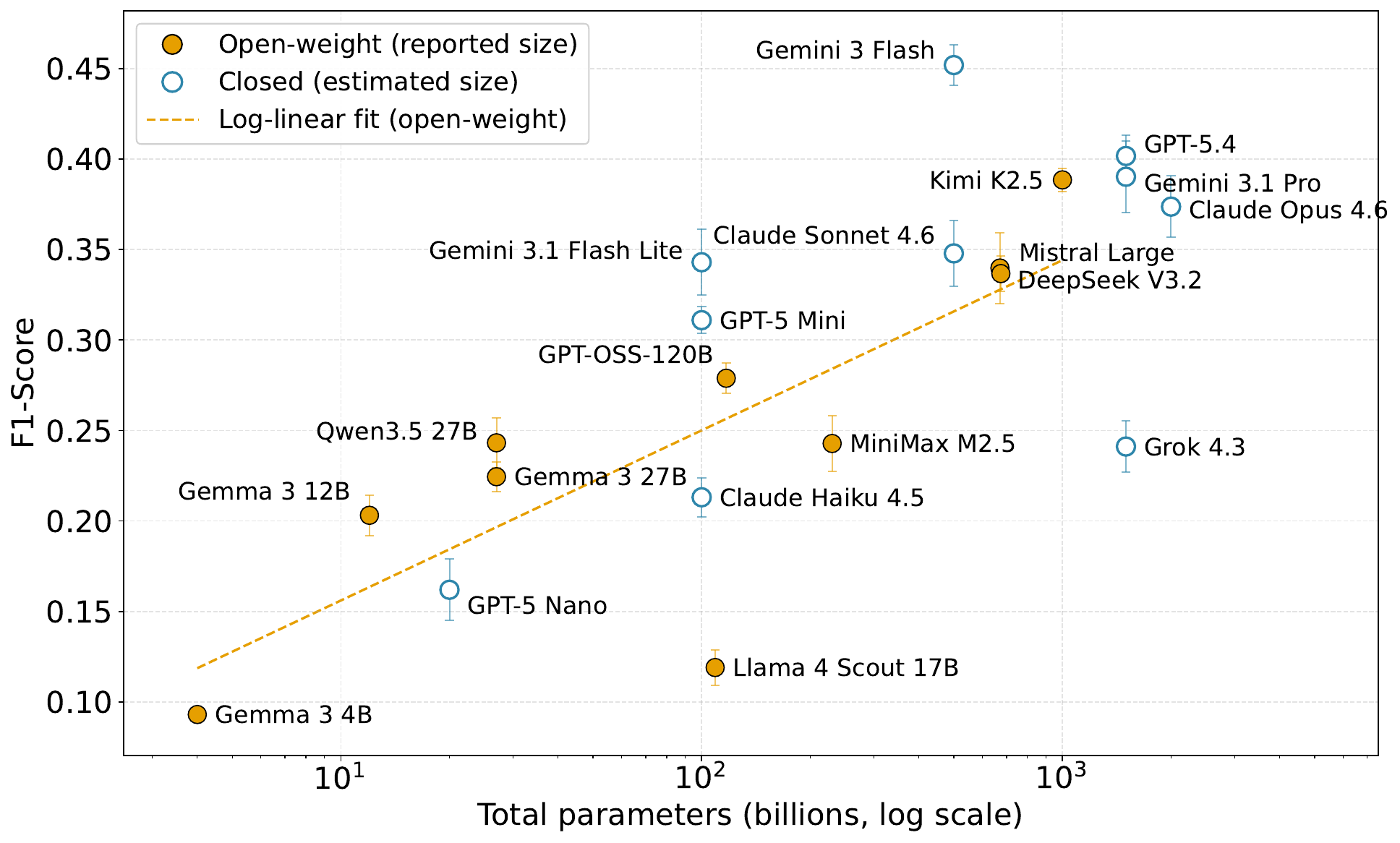}
    \caption{Experiment 1 - F1-Score (mean $\pm$ SD over 5 runs) vs.\ total parameter count. Filled: open-weight models (reported size); hollow: closed models (rough estimates, not used in the fit). Dashed: log-linear fit on open-weight models.}
    \label{fig:exp1_3}
\end{figure}

The precision-recall scatter plot (Figure \ref{fig:exp1_2}) reveals a clear frontier boundary: no model simultaneously achieves both high precision and high recall. Models fall along a curve where gains in one dimension tend to come at the cost of the other.

High-precision / low-recall models (GPT-5 Nano 92.8\% P / 8.9\% R, Grok 4.3 88.7 / 14.0, Claude Haiku 4.5 77.4 / 12.4) produce fewer but more precise triples, whereas Gemini 3 Flash reaches the highest recall (34.9\%) at moderate precision (64.0\%), which gives it the best F1; next come models combining high precision with comparatively high recall (GPT-5.4 84.4 / 26.4, Gemini 3.1 Pro 88.7 / 25.0). As no model exceeds 35\% recall while precision reaches 92.8\%, the limiting factor is not the precision of what models assert, but how much they fail to assert.

Interestingly, we observe a near-perfect horizontal separation on the y axis between blue and orange dots at precision around 73\%, meaning that, with Gemini 3 Flash being the only outlier, all closed models achieve higher precision than open-weight models.

\paragraph{Model scaling shows improvements.}F1 increases with size both within and across model families.

Within model families (shown with dotted arrows from small to large in Figure \ref{fig:exp1_2}): Anthropic Claude (0.213 → 0.348 → 0.374), OpenAI GPT (0.162 → 0.311 → 0.402), and Google Gemma (0.093 → 0.203 → 0.224). The biggest improvement in performance is observed from the smallest model to the medium model for all three families. Scaling affects precision and recall differently: larger Claude models gain recall while precision stays roughly constant, GPT-5.4 improves both over GPT-5 Mini (GPT-5 Nano is an outlier with 92.8\% precision but 8.9\% recall), and for Gemma recall grows monotonically while precision grows mainly from 12B to 27B.

Across model families (Figure \ref{fig:exp1_3}), for the ten open-weight models, whose sizes are reported by their developers, F1 correlates strongly with the logarithm of total parameter count (Pearson r = 0.81, p = 0.005; Spearman $\rho$ = 0.86, p = 0.002). Closed models do not disclose their size; plotting them at rough order-of-magnitude estimates yields a similar picture (Pearson r = 0.77, p $<$ 0.001; Spearman $\rho$ = 0.72, p $<$ 0.001). Size is however not the only factor: Llama 4 Scout (109B total parameters, F1 = 0.119) scores below the 12B Gemma 3 (0.203), and at equal size Qwen3.5 27B outperforms Gemma 3 27B (0.243 vs.\ 0.224).

We also find a strong positive correlation between F1 and output token price (Pearson r = 0.69, p $<$ 0.001; Spearman $\rho$ = 0.71, p $<$ 0.001), with more expensive models performing better but with several open-weight models offering a better quality-price ratio than closed ones (in addition to the possibility of being hosted locally)  e.g., DeepSeek V3.2 and GPT-OSS-120B reach F1 = 0.340 and 0.279 at \$0.42 and \$0.17, above GPT-5 Mini (0.311 at \$2.00) and Claude Haiku 4.5 (0.213 at \$5.00) (details in Appendix \ref{app:cost}).

\begin{figure*}[t]
    \centering
    \begin{subfigure}[b]{0.49\textwidth}\centering
        \includegraphics[width=\linewidth]{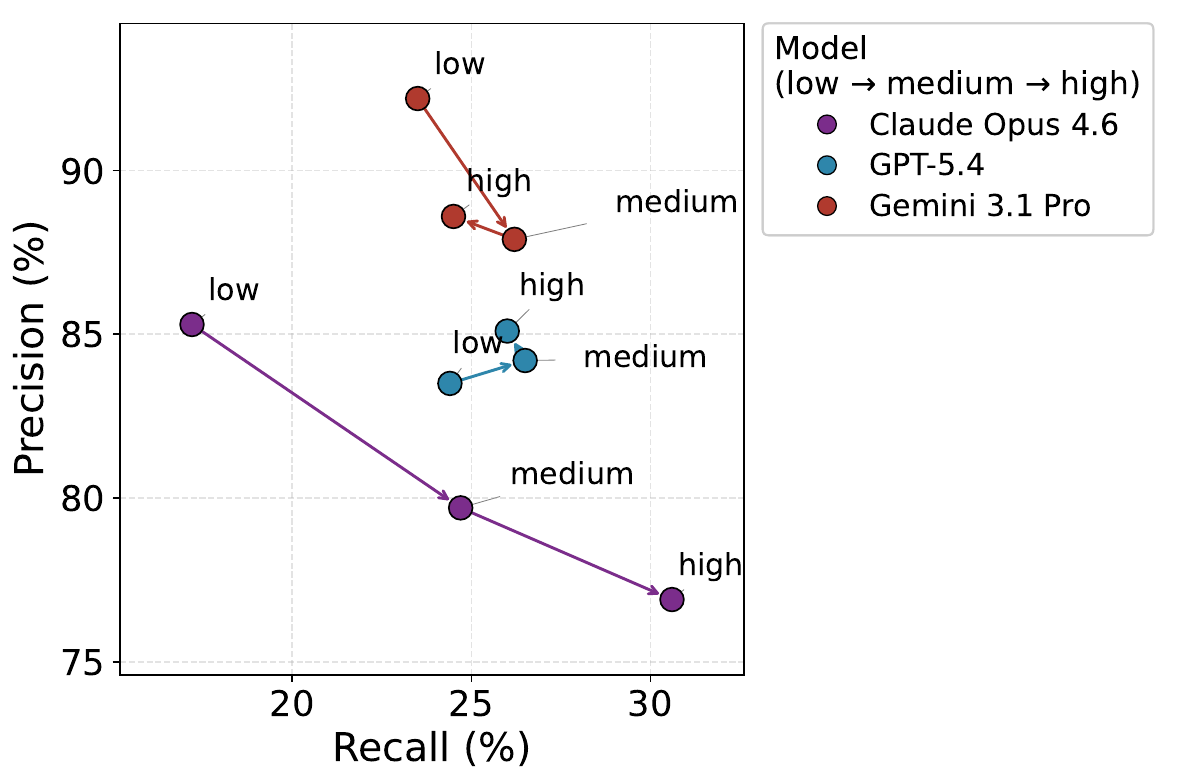}
        \caption{Experiment 2 - by reasoning effort.}\label{fig:exp2_1}
    \end{subfigure}\hfill
    \begin{subfigure}[b]{0.49\textwidth}\centering
        \includegraphics[width=\linewidth]{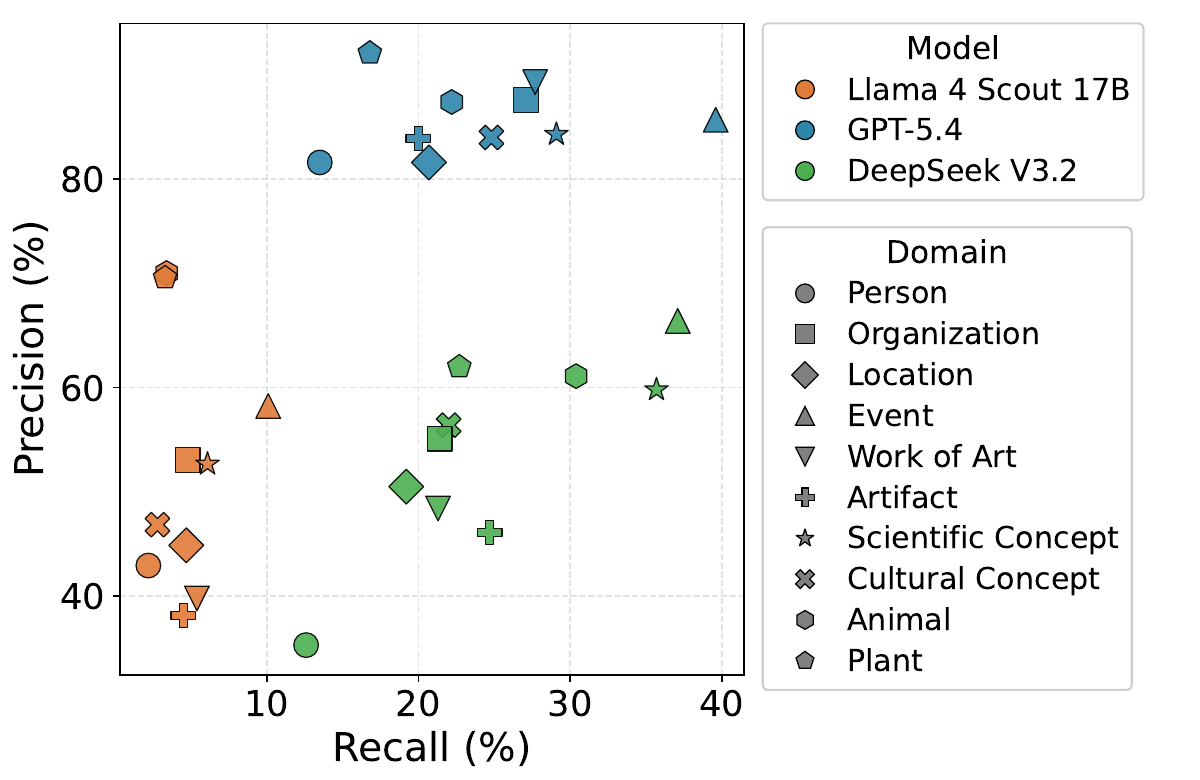}
        \caption{Experiment 3 - by domain.}\label{fig:exp3_1}
    \end{subfigure}
    \begin{subfigure}[b]{0.49\textwidth}\centering
        \includegraphics[width=\linewidth]{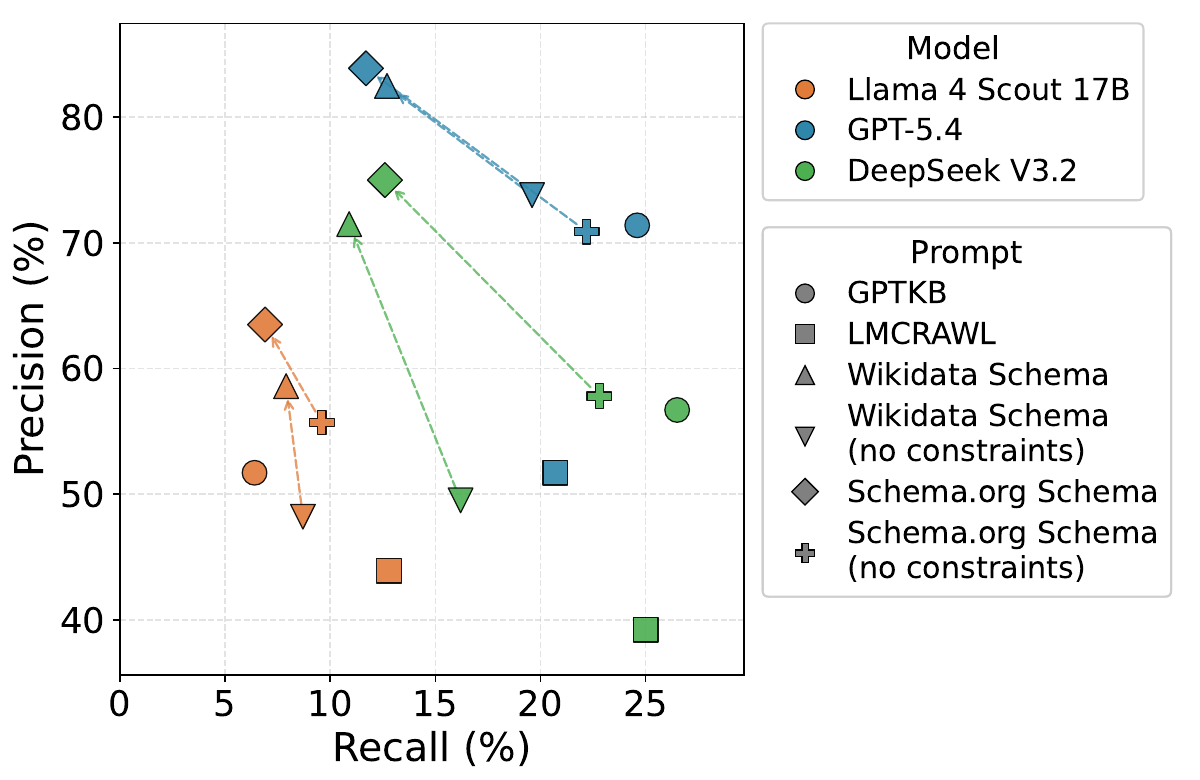}
        \caption{Experiment 4 - by prompt format.}\label{fig:exp5_1}
    \end{subfigure}\hfill
    \begin{subfigure}[b]{0.49\textwidth}\centering
        \includegraphics[width=\linewidth]{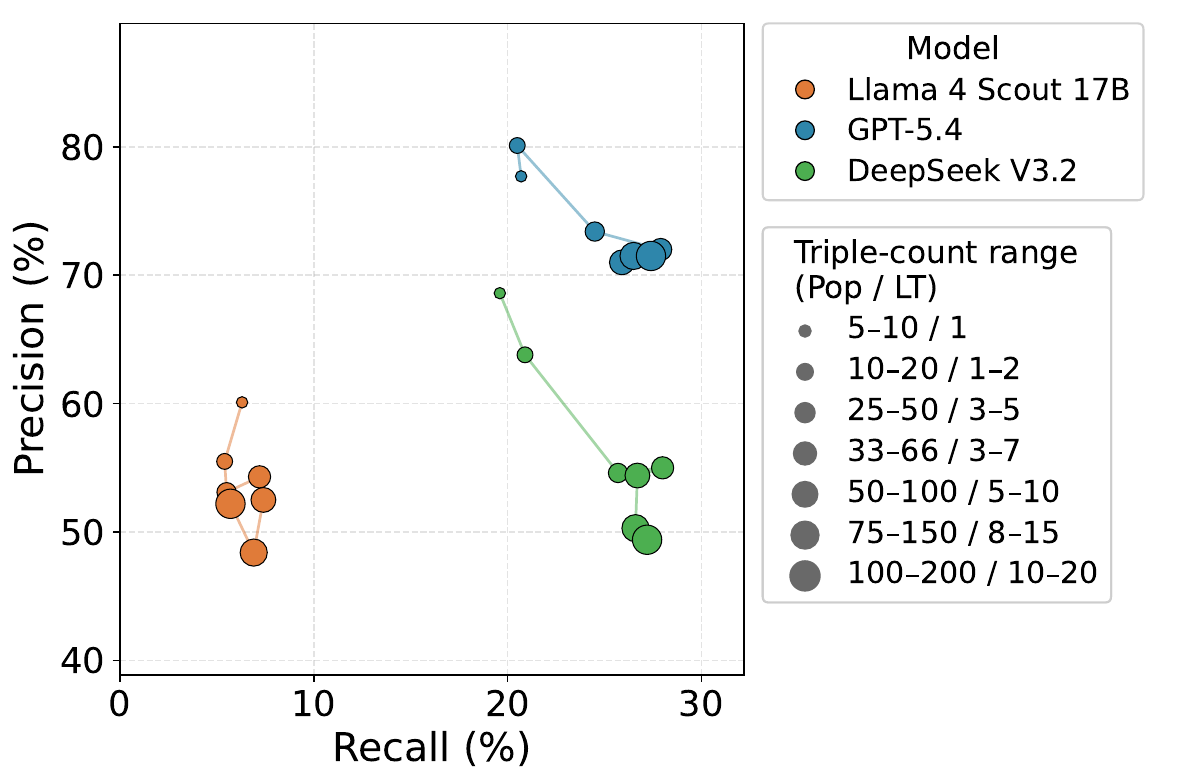}
        \caption{Experiment 5 - by triple range.}\label{fig:exp6_1}
    \end{subfigure}
    \caption{Precision-recall tradeoff in Experiments 2--5. In (c), dashed arrows go from the unconstrained to the predicate-constrained version of each schema.}
    \label{fig:exp2_5}
\end{figure*}
\subsection{Experiment 2: By reasoning effort}
Results: Figure \ref{fig:exp2_1} and Table \ref{app:full_results_exp2} (Appendix).

\paragraph{The impact of reasoning effort is model-dependent.} For Claude Opus 4.6, F1 increases monotonically with effort (0.286, 0.377, 0.438), driven entirely by recall (17.2\% → 30.6\%) while precision drops (85.3\% → 76.9\%). The outputs reveal the mechanism: at low effort, Opus returns a reasoning trace for only 60 of the 500 entities, leaves 90 entities empty (15 at high effort) and produces 11 instead of 23 triples per entity, so higher effort mainly increases its willingness to assert knowledge. For GPT-5.4 (0.378, 0.403, 0.398) and Gemini 3.1 Pro (0.374, 0.404, 0.384) the spread is at most 0.03 F1: reasoning effort is not a general lever for knowledge expression.

\subsection{Experiment 3: By domain + non-existent}
Results: Figure \ref{fig:exp3_1} and Table \ref{app:full_results_exp3} (Appendix).

\paragraph{Model ranking is coherent with Experiment 1.} GPT-5.4 outperforms DeepSeek V3.2 in seven of the ten domains (all but Animal, Plant and Scientific Concept), and DeepSeek V3.2 outperforms Llama 4 Scout in all ten, mirroring the Experiment 1 ranking.

\paragraph{Event and Scientific Concept are the strongest domains, Person the weakest.} Event is the top domain for all three models (GPT-5.4 F1 = 0.542; DeepSeek V3.2 F1 = 0.476; Llama 4 Scout F1 = 0.172), and Scientific Concept the second best (0.433, 0.447, 0.109).

Person is the lowest-scoring domain for all three models: GPT-5.4 (0.232), DeepSeek V3.2 (0.186), and Llama 4 Scout (0.042). The failure is driven primarily by recall rather than precision: GPT-5.4, for instance, answers 32 of the 100 Person entities with an empty list (at most 16 in any other domain), reaching 81.6\% precision but only 13.5\% recall, which suggests deliberate abstention on people rather than missing knowledge, possibly an artifact of safety post-training.

\paragraph{Non-existent entities.} No model consistently abstains on the non-existent entities, which would be the desired behavior, and the models differ widely: GPT-5.4 hallucinates 213 triples for 22 of the 100 entities, DeepSeek V3.2 1,944 triples for 43 entities, and Llama 4 Scout 617 triples for 91 entities.
 
\subsection{Experiment 4: By prompt format}
Experiment 4 evaluates GPT-5.4, DeepSeek V3.2, and Llama 4 Scout under six prompt variations on the first subset. Results: Figure \ref{fig:exp5_1} and Table \ref{app:full_results_exp4} (Appendix).

\paragraph{Open-ended prompts outperform schema-constrained prompts in F1.} Schemas consistently increase precision but reduce recall: Schema.org Schema yields each model's highest precision (GPT-5.4 83.9\%, DeepSeek V3.2 75.0\%, Llama 4 Scout 63.5\%) with among the lowest recall (11.7\%, 12.6\%, 6.9\%). The best F1 always comes from an open-ended prompt: GPTKB for GPT-5.4 (0.366) and DeepSeek V3.2 (0.361), LMCRAWL for Llama 4 Scout (0.198 vs.\ 0.114 with GPTKB).

\paragraph{Predefined predicate lists drive the schema tradeoff.} Comparing the constrained and unconstrained versions of each schema reveals that the predicate list is the dominant factor: removing it recovers substantial recall at modest precision cost. Models differ in how strictly they follow the list: GPT-5.4 keeps 98--99\% of its predicates within the allowed list, Llama 4 Scout 84--88\%, and DeepSeek V3.2 only 78\% (Wikidata) and 59\% (Schema.org).

\subsection{Experiment 5: By triple range}
This experiment evaluates GPT-5.4, DeepSeek V3.2, and Llama 4 Scout. We vary the two triple-count ranges stated in the GPTKB prompt, for popular (Pop) and less popular, long-tail (LT) subjects; the default prompt corresponds to Pop: 50–100 / LT: 5–10. Results: Figure \ref{fig:exp6_1} and Table \ref{app:full_results_exp5} (Appendix).
For GPT-5.4, F1 rises from 0.327 (smallest range) to 0.402 (Pop: 33–66 / LT: 3–7) and then plateaus, driven by recall (20.7\% → 27.4\%) while precision decreases (77.7\% → 71–73\%).
DeepSeek V3.2 follows the same pattern (F1 0.305 → 0.371, then 0.348–0.358), while its precision drops with larger ranges (68.6\% → 49.4–55.0\%);
Llama 4 Scout stays in a low-recall regime (5.4–7.4\%, F1 0.098–0.130).

\section{Conclusion}
We introduced open knowledge evaluation, which asks what models generate in response to open-ended elicitation prompts rather than a fixed set of queries, and instantiated it in BeQu, a benchmark of 10,000 entities with per-entity reference corpora from Wikipedia and the web, evaluating 20 models in five experimental scenarios. Our most notable findings are: (1) \textbf{Benchmark durability:} All models are far from saturating open-ended knowledge generation, and open-source models are competitive with commercial ones; (2) \textbf{Knowledge is ``unreasonable'':} Unlike for many other tasks, reasoning effort has little effect on open-ended knowledge expression for most models, with the notable exception of Claude Opus 4.6, whose recall grows with effort; (3) \textbf{Hard-wired operating points:} Explicit precision-recall steering is possible to a limited extent; (4) \textbf{Schema-driven tradeoff:} Enforcing a structured schema at inference time visibly drives the tradeoff towards precision.

We hope BeQu provides a useful complement to existing benchmarks and that open knowledge evaluation becomes a standard dimension of LLM knowledge assessment.

\section*{Limitations}


\paragraph{Definition of knowledge in LLMs.} The interpretation and operationalization of the human-centric concept of knowledge in LLMs remains controversial \cite{bender2021dangers,herrmann2025standards}. Recent work argues that belief-like representations can be studied under explicit adequacy criteria \cite{herrmann2025standards}, and mechanistic evidence supports the existence of belief-like internal states \cite{coronaMendozza2026beliefs}. Still, the question remains unsettled. Pragmatic evidence suggests that factual knowledge sometimes may be task-dependent \cite{elhelo2026taskSpecificKB}, that models can internally encode some information that is not faithfully reflected in their surface outputs \cite{orgad2025knowMore,gekhman2025insideOut}, or even that models express different knowledge on artificial benchmarking requests, versus organic ones \citep{needham2025large, chaudhary2025evaluation}.
A community consensus on practical definitions has yet to emerge; in this work we talk about LLM knowledge in \emph{behavioural} terms, i.e. by observing the output of a model. This represents one among several possible protocols to define factual knowledge in LLMs \citep{razniewski-etal-2026-epistemic}.

\paragraph{Evaluation of closed-source models.}
Our benchmark includes closed-source LLMs, which are known to undergo intransparent change, or even discontinuation. Our evaluation methodology entirely relies on open models and is therefore long-term reproducible; however, evaluation of prompted knowledge from commercial models does not come with such guarantees.

%
%
%
\bibliographystyle{splncs04}
\bibliography{refs}

@inproceedings{hu2025enabling,
  title={{Enabling {LLM} knowledge analysis via extensive materialization}},
  author={Hu, Yujia and Nguyen, Tuan-Phong and Ghosh, Shrestha and Razniewski, Simon},
  booktitle={ACL},
  year={2025},
 doi="https://doi.org/10.18653/v1/2025.acl-long.789"
}

@proceedings{fever-2026-1,
    title = "Proceedings of the Ninth Fact Extraction and {VER}ification Workshop ({FEVER})",
    editor = "Akhtar, Mubashara  and
      Aly, Rami  and
      Cao, Rui  and
      Christodoulopoulos, Christos  and
      Cocarascu, Oana  and
      Guo, Zhijiang  and
      Mittal, Arpit  and
      Schlichtkrull, Michael  and
      Thorne, James  and
      Vlachos, Andreas",
    year = "2026",
    url = "https://aclanthology.org/2026.fever-1.0/",
    doi = "10.18653/v1/2026.fever-1.0",
}

@inproceedings{song2024veriscore,
  title={VeriScore: Evaluating the factuality of verifiable claims in long-form text generation},
  author={Song, Yixiao and Kim, Yekyung and Iyyer, Mohit},
  booktitle={Findings of EMNLP},
  year={2024},
  url={https://doi.org/10.48550/arXiv.2406.19276}
}

@inproceedings{cohen2023crawling,
    title = {{Crawling The Internal Knowledge-Base of Language Models}},
    author = "Cohen, Roi  and
      Geva, Mor  and
      Berant, Jonathan  and
      Globerson, Amir",
    editor = "Vlachos, Andreas  and
      Augenstein, Isabelle",
    booktitle = "Findings of EACL",
    month = may,
    year = "2023",
    url = "https://aclanthology.org/2023.findings-eacl.139/",
    doi = "10.18653/v1/2023.findings-eacl.139"
}

@article{liang2022holistic,
  title={{Holistic Evaluation of Language Models}},
  author={Liang, Percy and Bommasani, Rishi and Lee, Tony and Tsipras, Dimitris and Soylu, Dilara and Yasunaga, Michihiro and Zhang, Yian and Narayanan, Deepak and Wu, Yuhuai and Kumar, Ananya and others},
  journal={arXiv preprint arXiv:2211.09110},
  year={2022},
  doi="https://doi.org/10.48550/arXiv.2211.09110"
}

@inproceedings{kiela-etal-2021-dynabench,
    title = {{Dynabench: Rethinking Benchmarking in NLP}},
    author = "Kiela, Douwe  and
      Bartolo, Max  and
      Nie, Yixin  and
      Kaushik, Divyansh  and
      Geiger, Atticus  and
      Wu, Zhengxuan  and
      Vidgen, Bertie  and
      Prasad, Grusha  and
      Singh, Amanpreet  and
      Ringshia, Pratik  and
      Ma, Zhiyi  and
      Thrush, Tristan  and
      Riedel, Sebastian  and
      Waseem, Zeerak  and
      Stenetorp, Pontus  and
      Jia, Robin  and
      Bansal, Mohit  and
      Potts, Christopher  and
      Williams, Adina",
    editor = "Toutanova, Kristina  and
      Rumshisky, Anna  and
      Zettlemoyer, Luke  and
      Hakkani-Tur, Dilek  and
      Beltagy, Iz  and
      Bethard, Steven  and
      Cotterell, Ryan  and
      Chakraborty, Tanmoy  and
      Zhou, Yichao",
    booktitle = "NAACL: Human Language Technologies",
    month = jun,
    year = "2021",
    url = "https://aclanthology.org/2021.naacl-main.324/",
    doi = "10.18653/v1/2021.naacl-main.324"
}

@inproceedings{jacovi-etal-2023-stop,
    title = {{Stop Uploading Test Data in Plain Text: Practical Strategies for Mitigating Data Contamination by Evaluation Benchmarks}},
    author = "Jacovi, Alon  and
      Caciularu, Avi  and
      Goldman, Omer  and
      Goldberg, Yoav",
    editor = "Bouamor, Houda  and
      Pino, Juan  and
      Bali, Kalika",
    booktitle = "EMNLP",
    month = dec,
    year = "2023",
    url = "https://aclanthology.org/2023.emnlp-main.308/",
    doi = "10.18653/v1/2023.emnlp-main.308"
}

@inproceedings{bowman-dahl-2021-will,
    title = {{What Will it Take to Fix Benchmarking in Natural Language Understanding}},
    author = "Bowman, Samuel R.  and
      Dahl, George",
    editor = "Toutanova, Kristina  and
      Rumshisky, Anna  and
      Zettlemoyer, Luke  and
      Hakkani-Tur, Dilek  and
      Beltagy, Iz  and
      Bethard, Steven  and
      Cotterell, Ryan  and
      Chakraborty, Tanmoy  and
      Zhou, Yichao",
    booktitle = "NAACL: Human Language Technologies",
    month = jun,
    year = "2021",
    url = "https://aclanthology.org/2021.naacl-main.385/",
    doi = "10.18653/v1/2021.naacl-main.385"
}

@inproceedings{giordano-razniewski-2026-foundations,
    title = "Foundations of {LLM} {K}nowledge {M}aterialization: {T}ermination, {R}eproducibility, {R}obustness",
    author = "Giordano, Luca  and
      Razniewski, Simon",
    editor = "Demberg, Vera  and
      Inui, Kentaro  and
      Marquez, Llu{\'i}s",
    booktitle = "Findings of {EACL}",
    month = mar,
    year = "2026",
    url = "https://aclanthology.org/2026.findings-eacl.113/",
    doi = "10.18653/v1/2026.findings-eacl.113",
    ISBN = "979-8-89176-386-9"
}

@article{kahnemann,
title = {Availability: A heuristic for judging frequency and probability},
journal = {Cognitive Psychology},
year = {1973},
issn = {0010-0285},
doi = {https://doi.org/10.1016/0010-0285(73)90033-9},
url = {https://www.sciencedirect.com/science/article/pii/0010028573900339},
author = {Amos Tversky and Daniel Kahneman},
}

@article{bubeck2023sparks,
  title={Sparks of {A}rtificial {G}eneral {I}ntelligence: Early experiments with {GPT}-4},
  author={Bubeck, S{\'e}bastien and Chandrasekaran, Varun and Eldan, Ronen and Gehrke, Johannes and Horvitz, Eric and Kamar, Ece and Lee, Peter and Lee, Yin Tat and Li, Yuanzhi and Lundberg, Scott and Nori, Harsha and Palangi, Hamid and Tulio Ribeiro, Marco and Zhang, Yi},
  journal={arXiv preprint arXiv:2303.12712},
  year={2023},
doi="https://doi.org/10.48550/arXiv.2303.12712"
}

@article{ji2023survey,
  title={Survey of hallucination in natural language generation},
  author={Ji, Ziwei and Lee, Nayeon and Frieske, Rita and Yu, Tiezheng and Su, Dan and Xu, Yan and Ishii, Etsuko and Bang, Ye Jin and Madotto, Andrea and Fung, Pascale},
  journal={ACM computing surveys},
  year={2023},
doi="https://doi.org/10.1145/3571730"
}

@inproceedings{berglund2024reversal,
  title={{T}he {R}eversal {C}urse: {LLM}s trained on {\textquotedblleft}A is B{\textquotedblright} fail to learn {\textquotedblleft}B is A{\textquotedblright}},
  author={Lukas Berglund and Meg Tong and Maximilian Kaufmann and Mikita Balesni and Asa Cooper Stickland and Tomasz Korbak and Owain Evans},
  booktitle={ICLR},
  year={2024},
  url={https://openreview.net/forum?id=GPKTIktA0k}
}

@article{srivastava2023beyond,
  title={{Beyond the Imitation Game: Quantifying and extrapolating the capabilities of language models}},
  author={{BIG-bench Team}},
  journal={Transactions on Machine Learning Research},
  issn={2835-8856},
  year={2023},
  url={https://openreview.net/forum?id=uyTL5Bvosj},
  note={}
}

@inproceedings{yao-etal-2026-medqa,
    title = {{{M}ed{QA}-{CS}: Objective Structured Clinical Examination ({OSCE})-Style Benchmark for Evaluating {LLM} Clinical Skills}},
    author = "Yao, Zonghai  and
      Zhang, Zihao  and
      Tang, Chaolong  and
      Bian, Xingyu  and
      Zhao, Youxia  and
      Yang, Zhichao  and
      Wang, Junda  and
      Zhou, Huixue  and
      Jang, Won Seok  and
      Ouyang, Feiyun  and
      Yu, Hong",
    editor = "Demberg, Vera  and
      Inui, Kentaro  and
      Marquez, Llu{\'i}s",
    booktitle = "EACL",
    month = mar,
    year = "2026",
    url = "https://aclanthology.org/2026.eacl-long.292/",
    doi = "10.18653/v1/2026.eacl-long.292",
    ISBN = "979-8-89176-380-7"
}

@inproceedings{guha2023legalbench,
  title={{Legalbench: A collaboratively built benchmark for measuring legal reasoning in large language models}},
  author={Guha, Neel and Nyarko, Julian and Ho, Daniel and R\'{e}, Christopher and Chilton, Adam and K, Aditya and Chohlas-Wood, Alex and Peters, Austin and Waldon, Brandon and Rockmore, Daniel and Zambrano, Diego and Talisman, Dmitry and Hoque, Enam and Surani, Faiz and Fagan, Frank and Sarfaty, Galit and Dickinson, Gregory and Porat, Haggai and Hegland, Jason and Wu, Jfessica and Nudell, Joe and Niklaus, Joel and Nay, John and Choi, Jonathan and Tobia, Kevin and Hagan, Margaret and Ma, Megan and Livermore, Michael and Rasumov-Rahe, Nikon and Holzenberger, Nils and Kolt, Noam and Henderson, Peter and Rehaag, Sean and Goel, Sharad and Gao, Shang and Williams, Spencer and Gandhi, Sunny and Zur, Tom and Iyer, Varun and Li, Zehua},
  booktitle={NeurIPS},
  year={2023},
  doi= {https://proceedings.neurips.cc/paper_files/paper/2023/file/89e44582fd28ddfea1ea4dcb0ebbf4b0-Paper-Datasets_and_Benchmarks.pdf}
}

@inproceedings{wei2024long,
  title={{Long-Form Factuality in Large Language Models}},
  author={Wei, Jerry and Yang, Chengrun and Song, Xinying and Lu, Yifeng and Hu, Nathan and Huang, Jie and Tran, Dustin and Peng, Daiyi and Liu, Ruibo and Huang, Da and Du, Cosmo and Le, Quoc V.},
  booktitle={NeurIPS},
  year={2024},
  doi= {https://dl.acm.org/doi/10.5555/3737916.3740483}
}

@article{center2026benchmark,
  title={A benchmark of expert-level academic questions to assess {AI} capabilities},
  author={Center for AI Safety, . and Scale AI, . and HLE Contributors Consortium, .},
  journal={Nature},
  year={2026},
  publisher={Nature Publishing Group UK London},
  url={https://www.nature.com/articles/s41586-025-09962-4}
}

@article{islam2023financebench,
  title={{FinanceBench: A New Benchmark for Financial Question Answering}},
  author={Islam, Pranab and Kannappan, Anand and Kiela, Douwe and Qian, Rebecca and Scherrer, Nino and Vidgen, Bertie},
  journal={arXiv preprint arXiv:2311.11944},
  year={2023},
  doi={https://doi.org/10.48550/arXiv.2311.11944}
}

@inproceedings{min-etal-2023-factscore,
    title = {{{FA}ct{S}core: Fine-grained Atomic Evaluation of Factual Precision in Long Form Text Generation}},
    author = "Min, Sewon  and
      Krishna, Kalpesh  and
      Lyu, Xinxi  and
      Lewis, Mike  and
      Yih, Wen-tau  and
      Koh, Pang  and
      Iyyer, Mohit  and
      Zettlemoyer, Luke  and
      Hajishirzi, Hannaneh",
    editor = "Bouamor, Houda  and
      Pino, Juan  and
      Bali, Kalika",
    booktitle = "EMNLP",
    month = dec,
    year = "2023",
    url = "https://aclanthology.org/2023.emnlp-main.741/",
    doi = "10.18653/v1/2023.emnlp-main.741"
}

@inproceedings{hu2026constructing,
  title={Constructing Disambiguated Knowledge Bases from Large Language Models at Scale},
  author={Hu, Yujia and Nguyen, Tuan-Phong and Razniewski, Simon},
  booktitle={AACL},
  year={2026},
  url={https://arxiv.org/pdf/2608.03729}
}

@inproceedings{wadhwa-etal-2023-revisiting,
    title = {{Revisiting Relation Extraction in the era of Large Language Models}},
    author = "Wadhwa, Somin  and
      Amir, Silvio  and
      Wallace, Byron",
    editor = "Rogers, Anna  and
      Boyd-Graber, Jordan  and
      Okazaki, Naoaki",
    booktitle = "ACL",
    year = "2023",
    url = "https://aclanthology.org/2023.acl-long.868/",
    doi = "10.18653/v1/2023.acl-long.868"
}

@inproceedings{hendrycks2020measuring,
  title={{Measuring Massive Multitask Language Understanding}},
  author={Hendrycks, Dan and Burns, Collin and Basart, Steven and Zou, Andy and Mazeika, Mantas and Song, Dawn and Steinhardt, Jacob},
  booktitle={ICLR},
  year={2021},
  url={https://openreview.net/pdf?id=d7KBjmI3GmQ}
}

@article{kwiatkowski-etal-2019-natural,
    title = {{Natural Questions: A Benchmark for Question Answering Research}},
    author = "Kwiatkowski, Tom  and
      Palomaki, Jennimaria  and
      Redfield, Olivia  and
      Collins, Michael  and
      Parikh, Ankur  and
      Alberti, Chris  and
      Epstein, Danielle  and
      Polosukhin, Illia  and
      Devlin, Jacob  and
      Lee, Kenton  and
      Toutanova, Kristina  and
      Jones, Llion  and
      Kelcey, Matthew  and
      Chang, Ming-Wei  and
      Dai, Andrew M.  and
      Uszkoreit, Jakob  and
      Le, Quoc  and
      Petrov, Slav",
    editor = "Lee, Lillian  and
      Johnson, Mark  and
      Roark, Brian  and
      Nenkova, Ani",
    journal = "Transactions of the Association for Computational Linguistics",
    year = "2019",
    url = "https://aclanthology.org/Q19-1026/",
    doi = "10.1162/tacl_a_00276"
}

@inproceedings{holtzman2020curious,
title={{The Curious Case of Neural Text Degeneration}},
author={Ari Holtzman and Jan Buys and Li Du and Maxwell Forbes and Yejin Choi},
booktitle={ICLR},
year={2020},
url={https://openreview.net/forum?id=rygGQyrFvH}
}

@article{saeed2026llmpedia,
  title={{LLMpedia: A Transparent Framework to Materialize an LLM's Encyclopedic Knowledge at Scale}},
  author={Saeed, Muhammed and Razniewski, Simon},
  journal={arXiv preprint arXiv:2603.24080},
  year={2026},
  doi={https://doi.org/10.48550/arXiv.2603.24080}
}

@inproceedings{bean2025measuring,
 author = {Bean, Andrew M. and Kearns, Ryan Othniel and Romanou, Angelika and Hafner, Franziska Sofia and Mayne, Harry and Batzner, Jan and Foroutan Eghlidi, Negar and Schmitz, Chris and Korgul, Karolina and Batra, Hunar and Deb, Oishi and Beharry, Emma and Emde, Cornelius and Foster, Thomas and Gausen, Anna and Grandury, Mar\'{\i}a and Han, Sophia and Hofmann, Valentin and Ibrahim, Lujain and Kim, Hazel and Kirk, Hannah Rose and Lin, Fangru and Liu, Gabrielle and Luettgau, Lennart and Magomere, Jabez and Rystr\o m, Jonathan and Sotnikova, Anna and Yang, Yushi and Zhao, Yilun and Bibi, Adel and Bosselut, Antoine and Clark, Ronald and Cohan, Arman and Foerster, Jakob and Gal, Yarin and Hale, Scott and Raji, Deborah and Summerfield, Christopher and Torr, Philip and Ududec, Cozmin and Rocher, Luc and Mahdi, Adam},
 booktitle = {NeurIPS},
 title = {Measuring what Matters: Construct Validity in Large Language Model Benchmarks},
 url = {https://proceedings.neurips.cc/paper_files/paper/2025/file/1967e0fc3aa6cbbace562f5cb8e3954e-Paper-Datasets_and_Benchmarks_Track.pdf},
 year = {2025}
}

@inproceedings{raji2021ai,
title={{AI} and the Everything in the Whole Wide World Benchmark},
author={Inioluwa Deborah Raji and Emily Denton and Emily M. Bender and Alex Hanna and Amandalynne Paullada},
booktitle={NeurIPS},
year={2021},
url={https://openreview.net/forum?id=j6NxpQbREA1}
}

@inproceedings{bender2021dangers,
  author = {Bender, Emily M. and Gebru, Timnit and McMillan-Major, Angelina and Shmitchell, Shmargaret},
  title = {On the Dangers of Stochastic Parrots: Can Language Models Be Too Big?},
  booktitle = {FAT},
  year = {2021},
  doi = {10.1145/3442188.3445922}
}

@article{herrmann2025standards,
  author = {Herrmann, Daniel A. and Levinstein, Benjamin A.},
  title = {Standards for Belief Representations in LLMs},
  journal = {Minds and Machines},
  year = {2025},
  doi = {10.1007/s11023-024-09709-6}
}

@misc{elhelo2026taskSpecificKB,
  author = {Elhelo, Amit and Globerson, Amir and Geva, Mor},
  title = {{LMs} as Task-Specific Knowledge Bases: An Interpretability Analysis},
  year = {2026},
  archivePrefix = {arXiv},
  doi = {10.48550/arXiv.2606.27237}
}

@article{orgad2025knowMore,
  author = {Orgad, Hadas and Toker, Michael and Gekhman, Zorik and Reichart, Roi and Szpektor, Idan and Kotek, Hadas and Belinkov, Yonatan},
  title = {{LLMs} Know More Than They Show: On the Intrinsic Representation of {LLM} Hallucinations},
  journal={ICLR},
  year = {2025},
  doi = {10.48550/arXiv.2410.02707}
}

@inproceedings{coronaMendozza2026beliefs,
  author = {Mendozza, Alessandro and S{\o}gaard, Anders},
  title = {{LLM} Beliefs Are in Their Heads},
  booktitle = {ACL},
  year = {2026},
  doi = {10.18653/v1/2026.acl-long.1905},
  url = {https://aclanthology.org/2026.acl-long.1905/}
}

@inproceedings{gekhman2025insideOut,
  author = {Gekhman, Zorik and Ben David, Eyal and Orgad, Hadas and Ofek, Eran and Belinkov, Yonatan and Szpektor, Idan and Herzig, Jonathan and Reichart, Roi},
  title = {Inside-Out: Hidden Factual Knowledge in {LLMs}},
  booktitle = {COLM},  
  year = {2025},
  doi = {10.48550/arXiv.2503.15299}
}

@article{needham2025large,
  title={{Large Language Models Often Know When They Are Being Evaluated}},
  author={Needham, Joe and Edkins, Giles and Pimpale, Govind and Bartsch, Henning and Hobbhahn, Marius},
  journal={arXiv},
  url={https://arxiv.org/abs/2505.23836},
  year={2025}
}

@article{chaudhary2025evaluation,
  title={{Evaluation Awareness Scales Predictably in Open-Weights Large Language Models}},
  author={Chaudhary, Maheep and Su, Ian and Hooda, Nikhil and Shankar, Nishith and Tan, Julia and Zhu, Kevin and Lagasse, Ryan and Sharma, Vasu and Panda, Ashwinee},
  journal={arXiv},
  url={https://arxiv.org/abs/2509.13333},
  year={2025}
}

@inproceedings{razniewski-etal-2026-epistemic,
    title = "Epistemic Twins: Enabling a Symbolic Science of Language Model Knowledge",
    author = "Razniewski, Simon  and
      Ghosh, Shrestha  and
      Giordano, Luca  and
      Hu, Yujia  and
      Kowalzik, Josua  and
      Nguyen, Tuan-Phong",
    booktitle = "AKBC@EMNLP",
    year = "2026",
    url = "https://aclanthology.org/2026.akbc-1.1/",
    doi = "10.18653/2026.akbc-1.1",
    ISBN = "979-8-89176-464-4"
}
\appendix
\section*{Appendix}

\section{Benchmark website}
\label{app:benchmark_website}

The BeQu benchmark is accompanied by a publicly accessible website ({\small\url{https://bequ.scads.ai/}}) providing a continuously updated leaderboard, browsable data (entity lists, reference corpora, elicited triples and per-triple judgments), and methodology documentation. The homepage (Figure \ref{fig:website}) displays the overall model ranking, with one tab per experiment (reasoning effort, domain, prompt format, triple range, and the additional experiment on popularity), and shows per-model precision, recall, and F1 scores computed on the primary random subset using the default GPTKB prompt with medium reasoning effort. Users can toggle between all-models and open-weights-only views, and search for specific models. The site also provides direct access to the paper, the GitHub repository, and the benchmark data. The leaderboard is designed to remain live beyond the submission of this paper, allowing newly released models to be evaluated and added to the ranking over time as the field evolves.

\section{Model Access and Cost}
\label{app:cost}

Between September and October 2026, we access all 20 models via OpenRouter\footnote{\href{https://openrouter.ai/}{https://openrouter.ai/}}. The entity lists and reference corpora were built once with a locally-hosted Llama 4 Scout and are released as fixed artifacts. The LLM judge (Gemma 4 26B-A4B-it), selected separately (Section \ref{section_evaluation}), is also hosted locally, while the passage embeddings for retrieval are computed with OpenAI's text-embedding-3-small via OpenRouter.

To conduct our experiments, including retries and test runs, we spend around \textbf{650\$} in OpenRouter credits and \textbf{55\$} for the Brave Search API, for a total amount of \textbf{705\$}. It must be noted that the amount spent on the Brave Search API is a one-time cost to build each entity's reference corpus.

\label{app:price}
\begin{figure}
    \centering
    \includegraphics[width=1\linewidth]{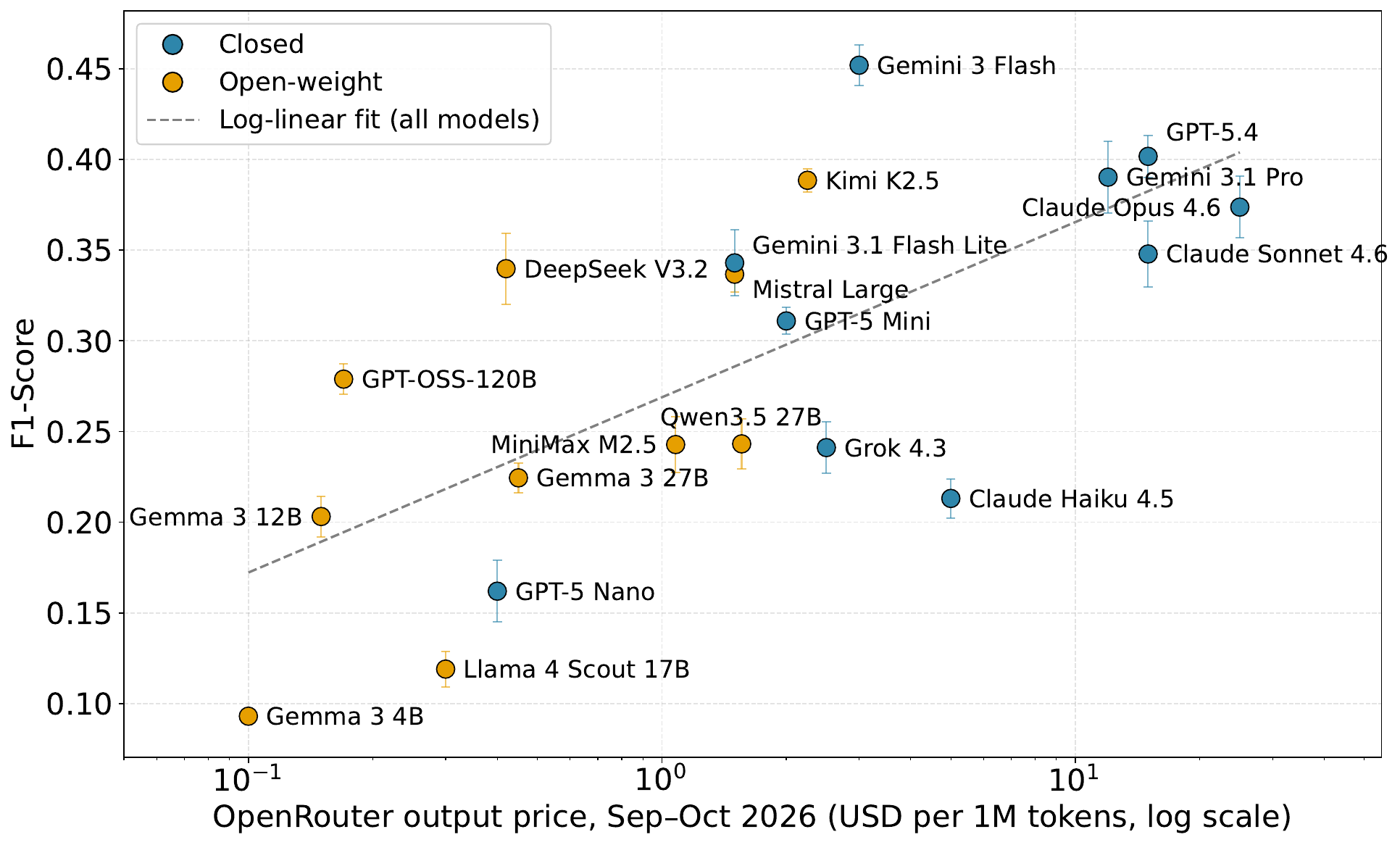}
    \caption{Experiment 1 - F1-Score (mean $\pm$ SD over 5 runs) vs.\ OpenRouter output price. Dashed: log-linear fit on all models.}
    \label{fig:price_f1}
\end{figure}
\paragraph{Price-F1 correlation.}We correlate each model's F1 in Experiment 1 with its OpenRouter output price per 1M tokens, taken as the median over weekly snapshots during the period of our experiments (September--October 2026). F1 grows with the logarithm of price (Pearson r = 0.69, p $<$ 0.001; Spearman $\rho$ = 0.71, p $<$ 0.001; Figure \ref{fig:price_f1}), for open-weight (r = 0.67, p = 0.04) and, not significantly, for closed models (r = 0.60, p = 0.07). Several open-weight models offer a better quality-price ratio than closed ones (in addition to the possibility of being hosted locally): DeepSeek V3.2 and GPT-OSS-120B reach F1 = 0.340 and 0.279 at \$0.42 and \$0.17, above GPT-5 Mini (0.311 at \$2.00) and Claude Haiku 4.5 (0.213 at \$5.00).

\section{Runtime}
\label{app:time}
Regarding time, it must be noted that 1) the time needed for building the entity lists strongly depends on the Wikipedia API backend, API backend/local hardware available used for the LLM judge, and the random extraction, 2) the time needed for the triple elicitation phase is strictly related to the API backend performance or local hardware available, 3) the time needed for building the reference corpus also strongly depends on the Wikipedia and Brave Search APIs, in addition to the API backend/local hardware available used for the LLM judge, and 4) the time needed for evaluation depends on the RAG retrieval speed, and the API backend/local hardware available used for the LLM judge. 

To build the entity lists, it required a magnitude of time of around 12h for each list, mainly driven by the large amount of unsuitable candidates. To build the reference corpora, it required around 2 minutes per entity. To evaluate the samples of elicited triples, it required on average 1 to 1.5 hours per setting (2,000 judgments), with several settings evaluated in parallel. Finally, the triple elicitation required between less than one and eight hours per setting, depending on the model and on the provider's load.

The main experiments were conducted between September and October 2026; Additional Experiments 1 and 2 were conducted in June 2026.

\section{Evaluation Example}
\label{app:eval_example}

\begin{figure*}
    \centering
    \includegraphics[width=1\linewidth]{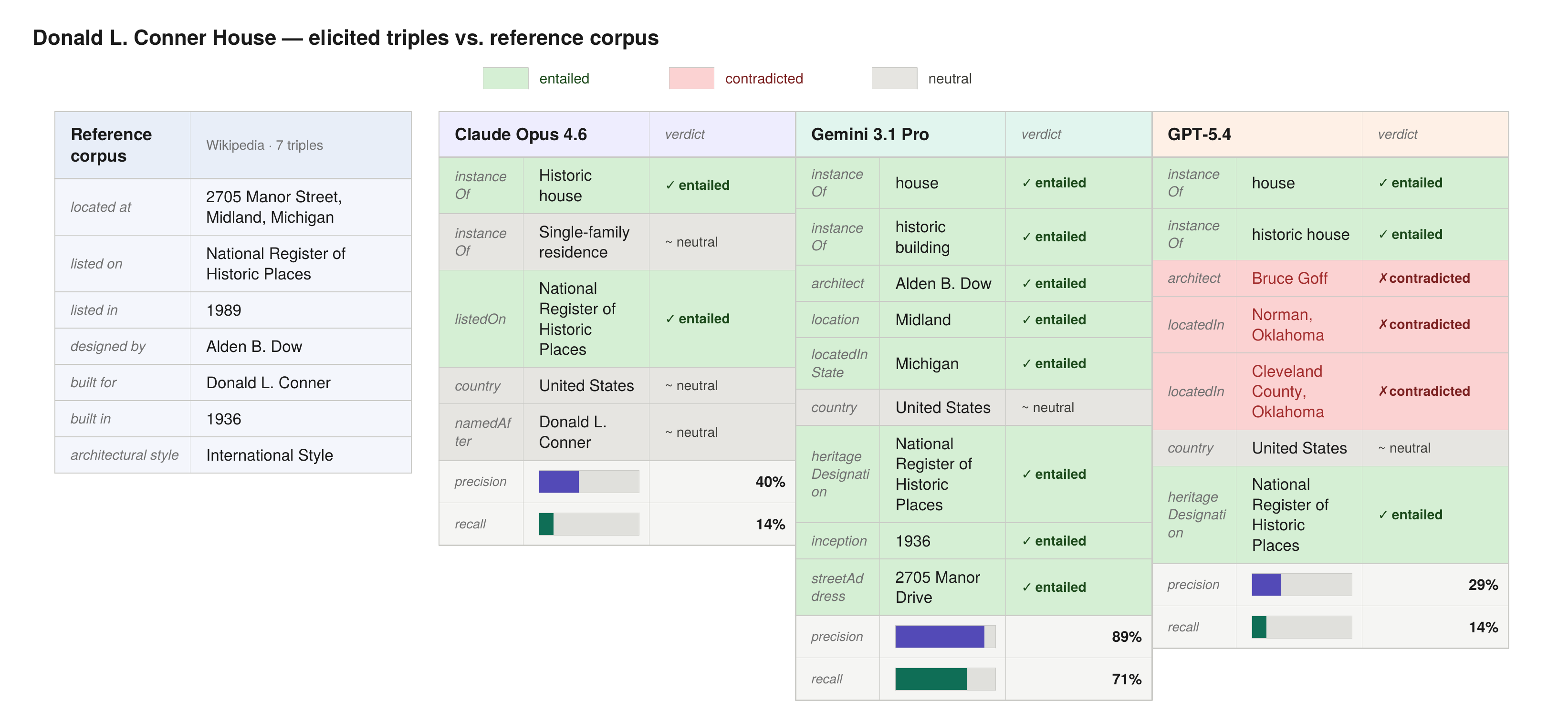}
    \caption{Illustrative example of open knowledge evaluation for the entity \textit{Donald L. Conner House}. Note: shown is just a sample of all triples actually generated.}
    \label{fig:example}
\end{figure*}

In Figure \ref{fig:example} we show an example of open knowledge evaluation on the entity \textit{Donald L. Conner House}, taken from a preliminary run of our pipeline. The leftmost column shows the reference triples extracted from Wikipedia, against which elicited triples are verified (for this illustration we ignore the triples extracted from the web). The three model columns show all triples elicited by Claude Opus 4.6, Gemini 3.1 Pro, and GPT-5.4 respectively, with their NLI verdict assigned by the Gemma 4 26B-A4B-it judge: entailed (green), contradicted (red), or neutral (grey). Precision and recall bars summarise aggregate performance for each model on this entity. Gemini 3.1 Pro achieves the highest recall, recovering the architect, address, construction year, and NRHP listing. Opus elicits only 5 triples, missing the architect, construction date, and address entirely. GPT-5.4 attributes the house to architect Bruce Goff and places it in Norman, Cleveland County, Oklahoma, all contradicted by the reference.

\section{Verbosity-F1 correlation analysis}
\label{app:verbosity}
We investigate the variable of verbosity by computing each model's average number of triples produced per entity in Experiment 1 and correlating it with its F1 (averaged over the five runs). We find no significant correlation (Pearson r = 0.20, p = 0.39; Spearman $\rho$ = 0.31, p = 0.18) across all 20 models. Hence, producing more triples per entity neither helps nor hurts a model's F1: verbose and concise models are found across the whole ranking.
 
\section{Additional Experiment 1 - Evolution Across LLM Family}

\begin{figure}
    \centering
    \includegraphics[width=1\linewidth]{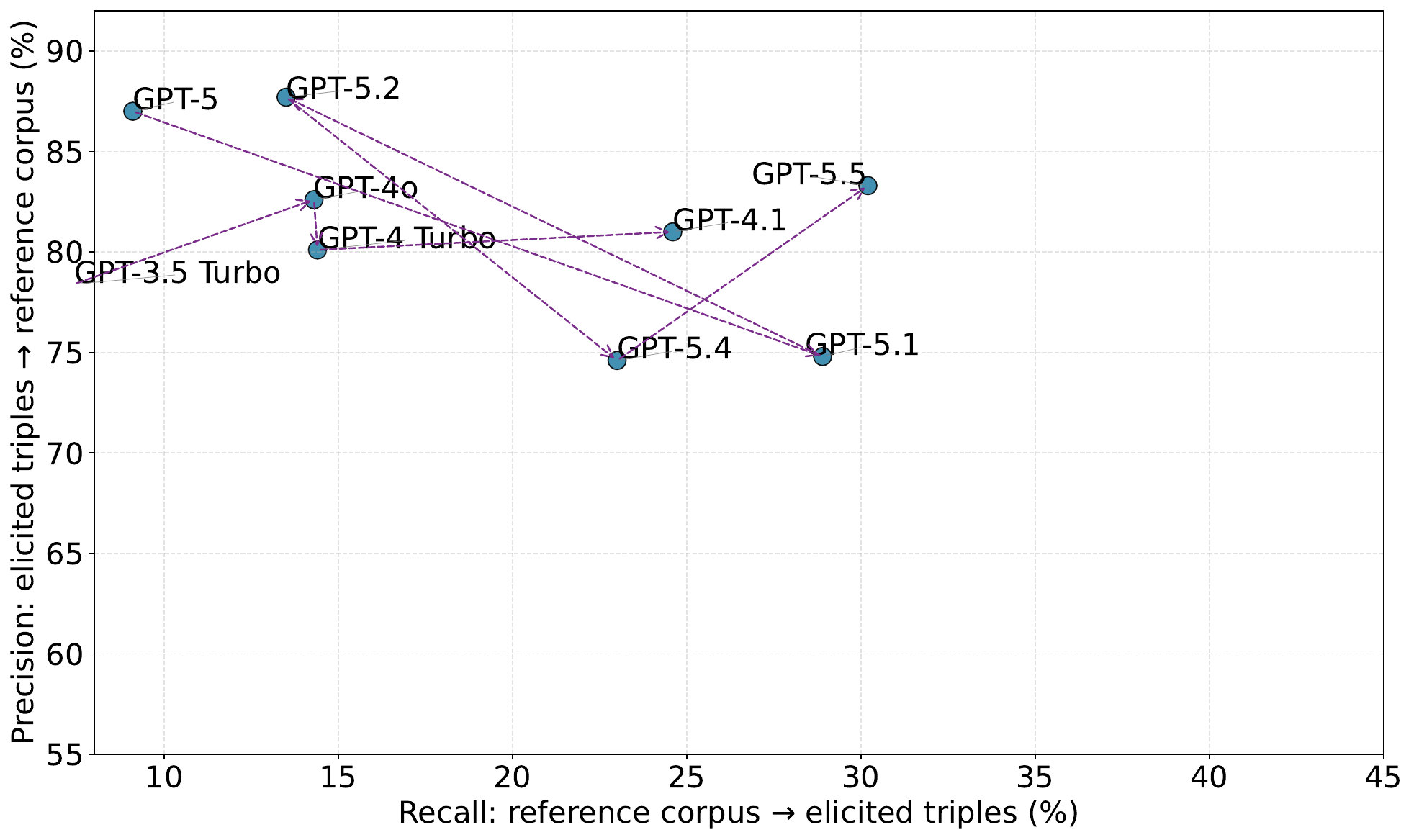}
    \caption{Additional Experiment 1 - Precision-recall tradeoff.}
    \label{fig:appE_fig1}
\end{figure}

\begin{figure}
    \centering
    \includegraphics[width=1\linewidth]{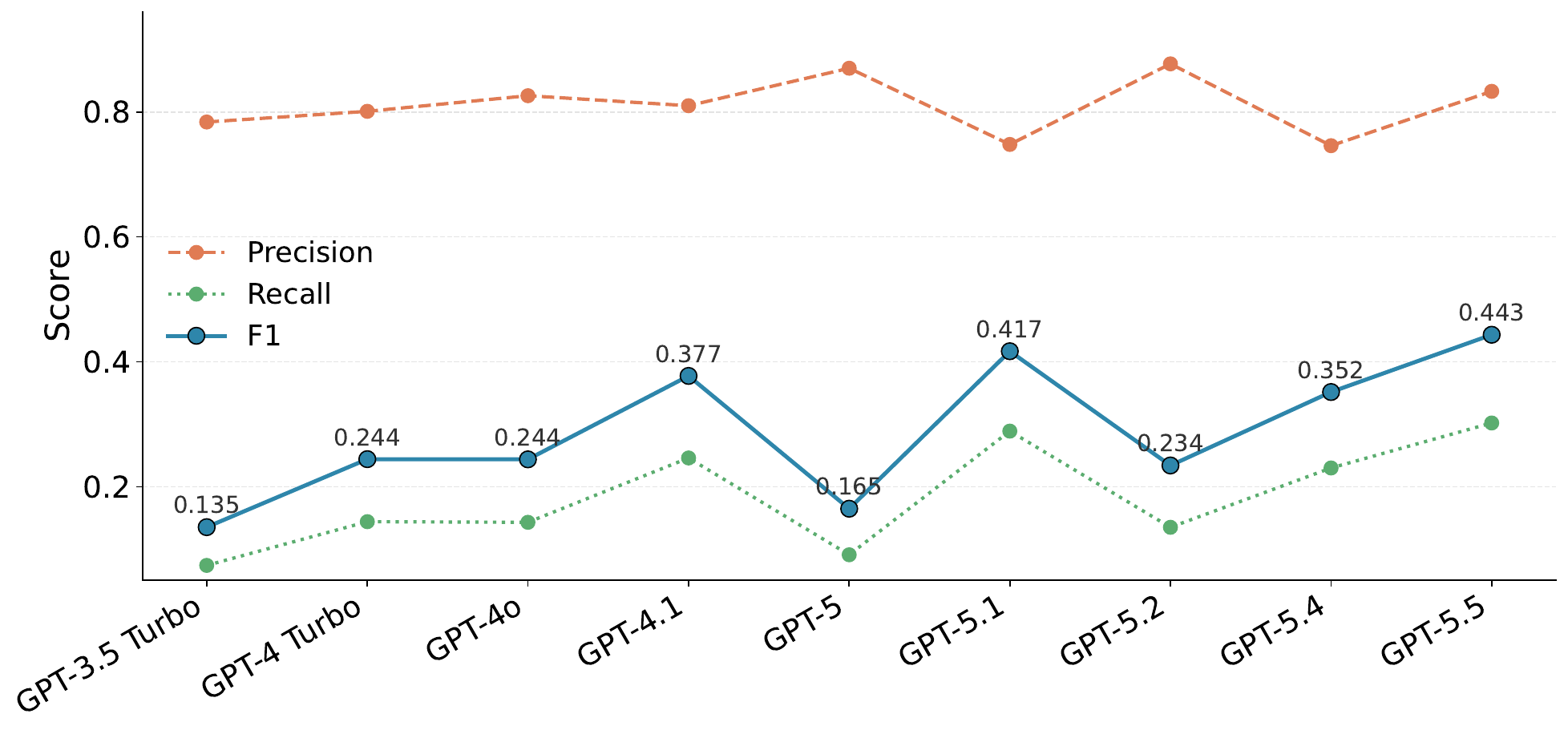}
    \caption{Additional Experiment 1 - Line chart.}
    \label{fig:appE_fig2}
\end{figure}
To investigate how factual knowledge expression has evolved across successive generations of an LLM family, we evaluate nine models from OpenAI's GPT family chronologically (from GPT-3.5 Turbo to GPT-5.5) on 200 entities of the primary random subset using the default GPTKB prompt. The answers of this experiment predate the main experiments, but they are evaluated with the same setup as the main experiments (Gemma 4 26B-A4B as the NLI judge, 1,000 sampled triples per direction). Results are shown in Figures \ref{fig:appE_fig1} and \ref{fig:appE_fig2}.

The overall trend in entailment F1 is upward across generations, rising from 0.135 (GPT-3.5 Turbo) to 0.443 (GPT-5.5), yet the trajectory is non-monotone. Two notable dips occur: GPT-5 registers the lowest F1 among the GPT-5 series (0.165), followed by a partial recovery in GPT-5.1 (0.417) and a secondary dip at GPT-5.2 (0.234), before a sustained climb through GPT-5.4 and GPT-5.5. As shown in Figure \ref{fig:appE_fig2}, these oscillations are primarily driven by recall rather than precision: the dips in F1 at GPT-5 and GPT-5.2 coincide with sharp drops in recall, while precision remains comparatively stable and high throughout (75–88\%), making recall the volatile and ultimately limiting factor, a finding consistent with the main results of Experiment 1.

The precision-recall scatter (Figure \ref{fig:appE_fig1}) further reveals that successive models do not move uniformly along the precision-recall frontier. Earlier models (GPT-3.5 Turbo, GPT-4o and 4 Turbo, GPT-5) cluster in the high-precision/low-recall region, while later models (GPT-5.1, GPT-5.4, GPT-5.5) shift rightward, trading modest precision for substantially higher recall. GPT-5.2 is a notable outlier, achieving the highest precision of any model in the family (87.7\%) while simultaneously posting the lowest recall among GPT-5.1 to GPT-5.5 (13.5\%), suggesting a generation-specific conservative output style. Taken together, these results indicate that progress in open knowledge expression across the GPT family is real but uneven, driven predominantly by improvements in recall, and that higher model generation does not guarantee a monotone improvement in F1.

\section{Additional Experiment 2 - By entity popularity}
\label{app:additional_exp_2}
We conducted an additional experiment to control for the effect of entity popularity on model performance. We constructed the \textbf{popularity subset} by partitioning the random subset in three buckets of exponentially-growing size (to mirror the typical pattern of entity popularity, where most entities are niche and a small minority are the most popular entities) sorted by descending entity popularity, representing high, mid, and low popularity. As a proxy for entity popularity we use the count of statements of each entity on Wikidata. We derive the two splitting points for the three buckets from the data distribution itself, and establish the first split at \textless20 statements (62\% of entities) and the second split at 20-49 statements (33\% of entities), leaving in the third bucket those with $\ge$50 statements, the top 5\% of entities. To keep the experiments tractable, we sample 200 random entities per popularity bucket, thus 600 entities in total.

\begin{figure}
    \centering
    \includegraphics[width=0.8\linewidth]{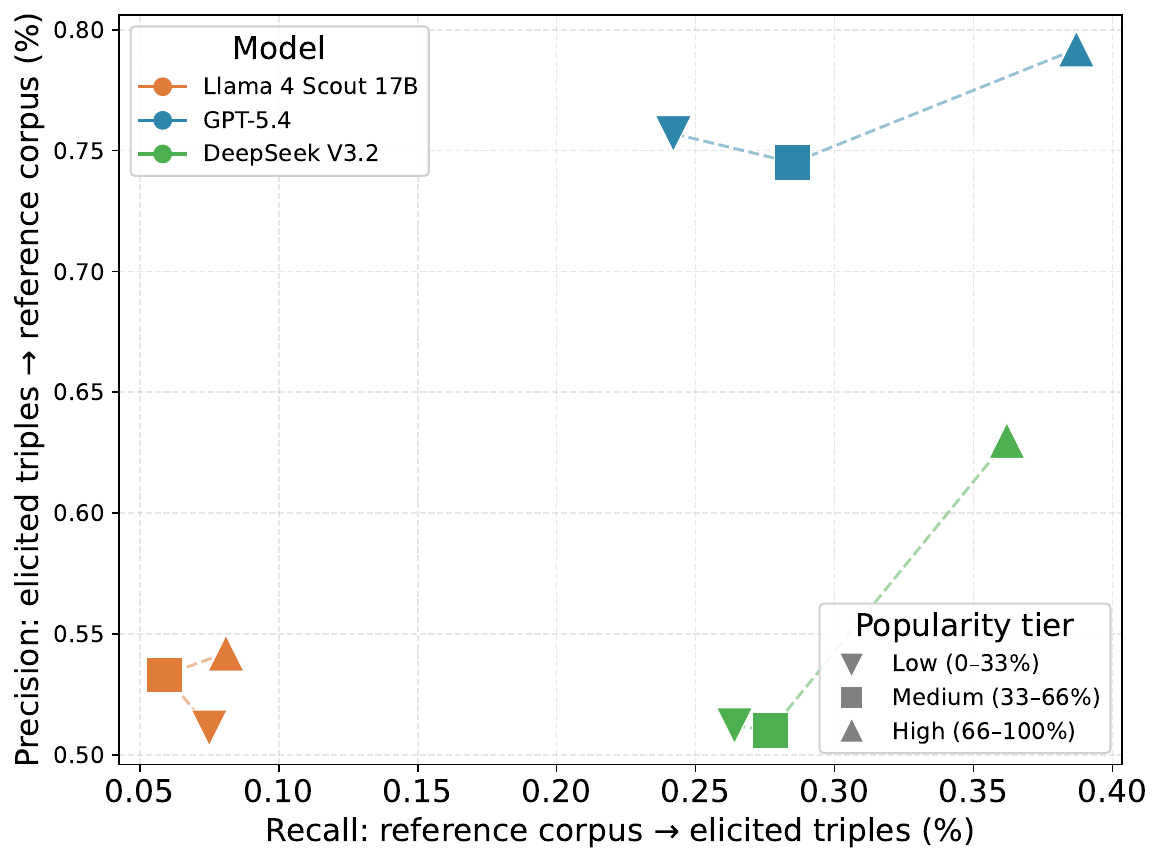}
    \caption{Precision-Recall tradeoff by entity popularity.}
    \label{fig:exp7_1}
\end{figure}
In this experiment we evaluate GPT-5.4, DeepSeek V3.2, and Llama 4 Scout. Like Additional Experiment 1, it is evaluated with the setup of the main experiments (Gemma 4 26B-A4B as the NLI judge, 1,000 sampled triples per direction and bucket). Results are shown in Figure \ref{fig:exp7_1}.

\paragraph{The impact of entity popularity is model-dependent.} For GPT-5.4, recall increases sharply with popularity (0.242, 0.285, 0.387), while precision stays high (0.757, 0.745, 0.792), shifting the operating point rightward. DeepSeek V3.2 shows a consistent trend in both directions: both precision and recall increase with popularity, mainly from mid to high (P: 0.512, 0.510, 0.630; R: 0.264, 0.277, 0.362). Llama 4 Scout remains in a low-recall regime across all tiers, with recall near-zero (0.075, 0.059, 0.081) and precision broadly stable (0.511, 0.533, 0.542), suggesting the model's behavior is largely unaffected by entity popularity.

\section{Error Analysis}
\label{app:error_analysis}

\subsection{Validation of the precision judge}
\label{app:validation_precision_judge}
To check the precision side of the evaluation, we submitted the reference triples themselves as elicited triples for the 500 entities of the random subset. Wikipedia reference triples (7,479) reach 91.5\% precision; all reference triples, Wikipedia and web (45,312), reach 64.1\% precision (91.2\% for Wikipedia, 56.2\% for web triples) and 92.9\% recall. As a negative control, Wikipedia triples whose object is replaced by the object of a triple with the same predicate of another entity (3,065 triples) reach only 2.2\% precision. The judge thus confirms nearly all facts stated in Wikipedia and almost never accepts false ones; the lower precision of web reference triples stems mostly from retrieval misses and from noisy or off-topic web triples.

\subsection{Evaluation}
\label{app:error_analysis_evaluation}

To better understand the sources of precision and recall loss, we conducted an error analysis of the NLI judge decisions across all 20 models in Experiment 1, focusing on the 116,491 triples labelled Neutral or Contradiction across the five runs (35,694 precision, 80,797 recall). We identify two broad categories: genuine model errors (found in Contradiction) and benchmark limitations (found in Neutral).

\paragraph{Model errors.} Precision contradictions are comparatively rare (11.5\% overall) but reveal systematic factual weaknesses. The most frequently contradicted predicates are generic class and location predicates (\texttt{instanceOf}, \texttt{locatedIn}), followed by temporal (\texttt{birthDate}, \texttt{dateOfBirth}, \texttt{deathDate}), geographical (\texttt{country}, \texttt{nationality}, \texttt{birthPlace}) and film attribution (\texttt{director}, \texttt{starring}). Representative examples include a model asserting that Robbie Ryan (cinematographer) was born in 1977 (Wikipedia records 1970), that Marie-Françoise Plissart is French rather than Belgian, and that Whitmell P. Martin was British rather than American. Error rates vary substantially across models, ranging from 3.1\% (GPT-5 Nano) to 25.3\% (Llama 4 Scout), with the smaller open-weight models (Gemma 3, Qwen3.5 27B) also among the most error-prone (13.3–19.4\%).

Recall contradictions are slightly less frequent (5.9\%). For example, one model elicits (\textit{Mikhail Pashnin}, \texttt{dateOfBirth}, \textit{1986-05-11}) while the reference records \textit{11 May 1989}; another assigns 20 February 1984 as the birth date of Jerome Messam, where the reference gives 2 April 1985. The Gemma family again stands out with recall contradiction rates of 7.5–14.2\%, together with GPT-OSS-120B (11.4\%) and Qwen3.5 27B (10.0\%).

\paragraph{Benchmark limitations.} The majority of non-Entailment decisions, however, are Neutral labels. Precision neutrals (24.2\%) arise in part because the top-10 retrieved passages do not always contain the elicited fact, even when it might be present elsewhere in the reference corpus: 19\% of neutral judge reasonings explicitly note that the fact is \emph{not explicitly} stated in the retrieved passages, as in the case of (\textit{Derozio Memorial College}, \texttt{locatedIn}, \textit{Rajarhat}), where the passages only state that the college is in Kolkata, of which Rajarhat is a locality.

Recall neutrals (74.9\%) are the largest single category overall; in most cases the elicited triples simply do not cover the reference fact, but 7\% of reasonings flag the fact as \emph{not explicitly} stated, showing that the judge sometimes requires a near-verbatim restatement of the reference triple, adopting a conservative, strict judgement behaviour. For instance, a model's output about Ramal de Braga describes it as a railway line and lists its connections, operator and length, but receives a Neutral label for the reference triple (\textit{Ramal de Braga}, \texttt{type}, \textit{branch line}) because no elicited triple states that it is a branch line.

Additionally, generic predicates such as \texttt{type}, \texttt{is a}, and \texttt{located in} appear prominently among recall-neutral cases.

These observations suggest two concrete directions for future improvement: expanding the passage retrieval window to reduce retrieval-induced false negatives on the precision side, and adopting a more lenient NLI criterion that awards partial credit for semantically equivalent expressions on the recall side.

\subsection{Reference corpus: coverage and temporal staleness}
\label{app:error_analysis_coverage_staleness}
To quantify the coverage of the reference corpus, we ran an experiment on 100 random entities, expanding their reference corpora from 20 to up to 200 web documents. Using GPT-OSS-120B as an LLM judge to classify whether triples newly extracted (by Llama 4 Scout) from each additional batch of documents are semantically novel relative to extractions so far, we find that 20 documents contain on average 55\% of the information contained in 50 documents, and 41\% of that contained in 200. Absolute recall figures should be interpreted relative to this measured document coverage. However, note that since every model in our experiments is evaluated against the same reference corpus, the comparative rankings remain robust.

Additionally, we investigate knowledge temporal staleness by sampling 100 random entities and classifying them into four categories: 1) knowledge about the entity will never change, 2) knowledge about the entity is unlikely to change, 3) knowledge about the entity is likely to change, even in short or medium term, 4) knowledge about the entity will certainly change, potentially a lot, and even in short term. 46\% of entities fall into the first category (historical events, deceased persons, chemical compounds etc.) and 29\% into the second (small localities, infrastructures, established institutions, retired but still alive public figures), with only 25\% classified into the third (young and alive public figures) and no entities classified into the fourth category. So temporal evolution conflicts are possible, but unlikely to dominate in the near future, and one could always subsample enough static entities.

\subsection{Entity selection}
\label{app:error_analysis_entity_selection}

We validate the entity selection step performed by Llama 4 Scout with a human evaluation and find an Average Mean Absolute Error of 0.96 for Informativeness, 0.96 for Ambiguity, and 0.82 for Suitability, for an overall average misalignment of 0.913 on a 1-5 scale. Although the LLM judge rates systematically higher almost 1 point than our human evaluation, we consider this an advantage for this task rather than a limitation: an LLM trained on huge amounts of data probably knows (at least something) about more entities than an average human, which would explain higher scores on Informativeness (“I know something about this entity”), and consequently on Ambiguity and Suitability. This allows for a larger pool of suitable long-tail entities to be selected.

\begin{figure*}
    \centering
    \includegraphics[width=1\linewidth]{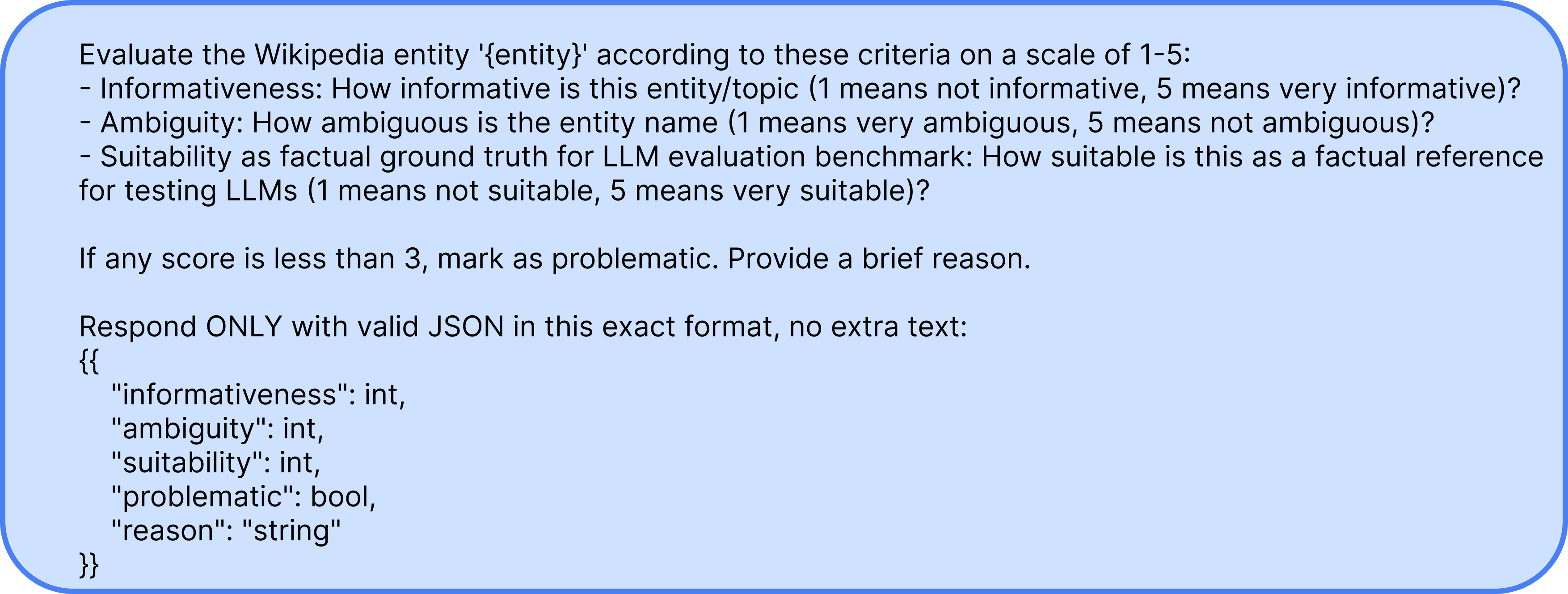}
    \caption{Prompt used for the LLM judge in Part 1 - Entity Selection. The \{entity\} variable is filled with the Wikipedia article's title.}
    \label{fig:judge_prompt_entities}
\end{figure*}

\section{Full results and statistics}
\label{app:full_results}

Full results of the experiments are in the Tables \ref{app:full_results_exp1}-\ref{app:full_results_exp5}. Statistics are in the tables \ref{app:elicit_stats_exp1}-\ref{app:elicit_stats_exp5}.

\begin{table}[!htbp]
\centering
\renewcommand{\arraystretch}{0.9}
\setlength{\tabcolsep}{2pt}
\scriptsize
\begin{adjustbox}{max width=\linewidth}
\begin{tabular}{lccccccc}
\toprule
\rowcolor{headerblue}
\textcolor{white}{\textbf{Model}} &
\textcolor{white}{\textbf{F1 (Ent.)}} &
\multicolumn{3}{c}{\textcolor{white}{\textbf{Precision Exp.}}} &
\multicolumn{3}{c}{\textcolor{white}{\textbf{Recall Exp.}}} \\
\cmidrule(lr){3-5}\cmidrule(lr){6-8}
\rowcolor{headerblue}
& &
\textcolor{white}{\textbf{E}} &
\textcolor{white}{\textbf{N}} &
\textcolor{white}{\textbf{C}} &
\textcolor{white}{\textbf{E}} &
\textcolor{white}{\textbf{N}} &
\textcolor{white}{\textbf{C}} \\
\midrule
\rowcolor{rowgray}
gemini-3-flash-preview & \ms{0.452}{.011} & \ms{0.640}{.016} & \ms{0.272}{.013} & \ms{0.088}{.011} & \ms{0.349}{.012} & \ms{0.584}{.011} & \ms{0.067}{.003} \\
gpt-5.4                & \ms{0.402}{.012} & \ms{0.844}{.006} & \ms{0.089}{.005} & \ms{0.067}{.002} & \ms{0.264}{.009} & \ms{0.699}{.014} & \ms{0.037}{.009} \\
\rowcolor{rowgray}
gemini-3.1-pro-preview & \ms{0.390}{.020} & \ms{0.887}{.012} & \ms{0.069}{.009} & \ms{0.044}{.006} & \ms{0.250}{.017} & \ms{0.721}{.020} & \ms{0.029}{.003} \\
kimi-k2.5              & \ms{0.388}{.006} & \ms{0.702}{.042} & \ms{0.199}{.035} & \ms{0.099}{.012} & \ms{0.269}{.004} & \ms{0.676}{.015} & \ms{0.055}{.013} \\
\rowcolor{rowgray}
claude-opus-4.6        & \ms{0.374}{.017} & \ms{0.791}{.011} & \ms{0.122}{.009} & \ms{0.087}{.007} & \ms{0.245}{.014} & \ms{0.710}{.014} & \ms{0.045}{.004} \\
claude-sonnet-4.6      & \ms{0.348}{.018} & \ms{0.767}{.006} & \ms{0.139}{.012} & \ms{0.095}{.008} & \ms{0.225}{.016} & \ms{0.734}{.019} & \ms{0.041}{.005} \\
\rowcolor{rowgray}
gemini-3.1-flash-lite  & \ms{0.343}{.018} & \ms{0.755}{.005} & \ms{0.154}{.005} & \ms{0.091}{.005} & \ms{0.222}{.015} & \ms{0.733}{.013} & \ms{0.045}{.009} \\
deepseek-v3.2          & \ms{0.340}{.020} & \ms{0.551}{.011} & \ms{0.305}{.011} & \ms{0.144}{.010} & \ms{0.246}{.020} & \ms{0.678}{.022} & \ms{0.076}{.011} \\
\rowcolor{rowgray}
mistral-large-2512     & \ms{0.337}{.010} & \ms{0.638}{.010} & \ms{0.215}{.012} & \ms{0.147}{.006} & \ms{0.229}{.009} & \ms{0.705}{.007} & \ms{0.066}{.007} \\
gpt-5-mini             & \ms{0.311}{.007} & \ms{0.823}{.015} & \ms{0.146}{.010} & \ms{0.031}{.006} & \ms{0.192}{.006} & \ms{0.794}{.007} & \ms{0.014}{.001} \\
\rowcolor{rowgray}
gpt-oss-120b           & \ms{0.279}{.008} & \ms{0.359}{.008} & \ms{0.475}{.019} & \ms{0.166}{.016} & \ms{0.228}{.012} & \ms{0.658}{.021} & \ms{0.114}{.012} \\
qwen3.5-27b            & \ms{0.243}{.014} & \ms{0.459}{.018} & \ms{0.366}{.024} & \ms{0.176}{.009} & \ms{0.166}{.012} & \ms{0.734}{.023} & \ms{0.100}{.012} \\
\rowcolor{rowgray}
minimax-m2.5           & \ms{0.243}{.015} & \ms{0.649}{.014} & \ms{0.210}{.016} & \ms{0.141}{.016} & \ms{0.149}{.011} & \ms{0.805}{.015} & \ms{0.046}{.007} \\
grok-4.3               & \ms{0.241}{.014} & \ms{0.887}{.010} & \ms{0.064}{.007} & \ms{0.049}{.012} & \ms{0.140}{.010} & \ms{0.844}{.009} & \ms{0.016}{.005} \\
\rowcolor{rowgray}
gemma-3-27b-it         & \ms{0.224}{.008} & \ms{0.335}{.024} & \ms{0.472}{.012} & \ms{0.194}{.019} & \ms{0.169}{.004} & \ms{0.710}{.007} & \ms{0.121}{.007} \\
claude-haiku-4.5       & \ms{0.213}{.011} & \ms{0.774}{.003} & \ms{0.135}{.008} & \ms{0.091}{.010} & \ms{0.124}{.007} & \ms{0.851}{.007} & \ms{0.026}{.003} \\
\rowcolor{rowgray}
gemma-3-12b-it         & \ms{0.203}{.011} & \ms{0.277}{.016} & \ms{0.589}{.039} & \ms{0.133}{.024} & \ms{0.161}{.015} & \ms{0.697}{.010} & \ms{0.142}{.012} \\
gpt-5-nano             & \ms{0.162}{.017} & \ms{0.928}{.008} & \ms{0.042}{.002} & \ms{0.031}{.008} & \ms{0.089}{.010} & \ms{0.902}{.008} & \ms{0.009}{.004} \\
\rowcolor{rowgray}
llama-4-scout          & \ms{0.119}{.010} & \ms{0.521}{.013} & \ms{0.226}{.016} & \ms{0.253}{.016} & \ms{0.067}{.006} & \ms{0.884}{.008} & \ms{0.049}{.005} \\
gemma-3-4b-it          & \ms{0.093}{.004} & \ms{0.276}{.021} & \ms{0.543}{.031} & \ms{0.181}{.017} & \ms{0.056}{.002} & \ms{0.869}{.004} & \ms{0.075}{.002} \\
\bottomrule
\end{tabular}
\end{adjustbox}
\caption{Full results for Experiment 1: mean $\pm$ standard deviation (subscript) over 5 independent runs. Sorted by mean F1 Score in descending order.}
\label{app:full_results_exp1}
\end{table}

\begin{table}[!htbp]
\centering
\renewcommand{\arraystretch}{0.9}
\setlength{\tabcolsep}{2pt}
\scriptsize
\begin{adjustbox}{max width=\linewidth}
\begin{tabular}{l l S[table-format=1.3] S[table-format=1.3] S[table-format=1.3] S[table-format=1.3] S[table-format=1.3] S[table-format=1.3] S[table-format=1.3]}
\toprule
\rowcolor{headerblue}
\textcolor{white}{\textbf{Model}} &
\textcolor{white}{\textbf{Reasoning}} &
{\textcolor{white}{\textbf{F1 (Ent.)}}} &
\multicolumn{3}{c}{\textcolor{white}{\textbf{Precision Exp.}}} &
\multicolumn{3}{c}{\textcolor{white}{\textbf{Recall Exp.}}} \\
\cmidrule(lr){4-6}\cmidrule(lr){7-9}
\rowcolor{headerblue}
& & &
{\textcolor{white}{\textbf{E}}} &
{\textcolor{white}{\textbf{N}}} &
{\textcolor{white}{\textbf{C}}} &
{\textcolor{white}{\textbf{E}}} &
{\textcolor{white}{\textbf{N}}} &
{\textcolor{white}{\textbf{C}}} \\
\midrule
\rowcolor{rowgray}
claude-opus-4.6        & high   & 0.438 & 0.769 & 0.141 & 0.090 & 0.306 & 0.648 & 0.046 \\
gemini-3.1-pro-preview & medium & 0.404 & 0.879 & 0.074 & 0.047 & 0.262 & 0.708 & 0.030 \\
\rowcolor{rowgray}
gpt-5.4                & medium & 0.403 & 0.842 & 0.092 & 0.066 & 0.265 & 0.688 & 0.047 \\
gpt-5.4                & high   & 0.398 & 0.851 & 0.094 & 0.055 & 0.260 & 0.707 & 0.033 \\
\rowcolor{rowgray}
gemini-3.1-pro-preview & high   & 0.384 & 0.886 & 0.075 & 0.039 & 0.245 & 0.725 & 0.030 \\
gpt-5.4                & low    & 0.378 & 0.835 & 0.098 & 0.067 & 0.244 & 0.711 & 0.045 \\
\rowcolor{rowgray}
claude-opus-4.6        & medium & 0.377 & 0.797 & 0.122 & 0.081 & 0.247 & 0.709 & 0.044 \\
gemini-3.1-pro-preview & low    & 0.374 & 0.922 & 0.041 & 0.037 & 0.235 & 0.733 & 0.032 \\
\rowcolor{rowgray}
claude-opus-4.6        & low    & 0.286 & 0.853 & 0.080 & 0.067 & 0.172 & 0.793 & 0.035 \\
\bottomrule
\end{tabular}
\end{adjustbox}
\caption{Full results for Experiment 2. Sorted by F1 Score in descending order.}
\label{app:full_results_exp2}
\end{table}

\begin{table}[!htbp]
\centering
\renewcommand{\arraystretch}{0.9}
\setlength{\tabcolsep}{2pt}
\scriptsize
\begin{adjustbox}{max width=\linewidth}
\begin{tabular}{l l S[table-format=1.3] S[table-format=1.3] S[table-format=1.3] S[table-format=1.3] S[table-format=1.3] S[table-format=1.3] S[table-format=1.3]}
\toprule
\rowcolor{headerblue}
\textcolor{white}{\textbf{Model}} &
\textcolor{white}{\textbf{Domain}} &
{\textcolor{white}{\textbf{F1 (Ent.)}}} &
\multicolumn{3}{c}{\textcolor{white}{\textbf{Precision Exp.}}} &
\multicolumn{3}{c}{\textcolor{white}{\textbf{Recall Exp.}}} \\
\cmidrule(lr){4-6}\cmidrule(lr){7-9}
\rowcolor{headerblue}
& & &
{\textcolor{white}{\textbf{E}}} &
{\textcolor{white}{\textbf{N}}} &
{\textcolor{white}{\textbf{C}}} &
{\textcolor{white}{\textbf{E}}} &
{\textcolor{white}{\textbf{N}}} &
{\textcolor{white}{\textbf{C}}} \\
\midrule
\rowcolor{rowgray}
deepseek-v3.2 & event              & 0.476 & 0.664 & 0.213 & 0.123 & 0.371 & 0.528 & 0.101 \\
              & scientific concept & 0.447 & 0.598 & 0.285 & 0.117 & 0.357 & 0.570 & 0.073 \\
\rowcolor{rowgray}
              & animal             & 0.406 & 0.611 & 0.295 & 0.094 & 0.304 & 0.637 & 0.059 \\
              & plant              & 0.332 & 0.620 & 0.262 & 0.118 & 0.227 & 0.721 & 0.052 \\
\rowcolor{rowgray}
              & artifact           & 0.322 & 0.461 & 0.332 & 0.207 & 0.247 & 0.611 & 0.142 \\
              & cultural concept   & 0.317 & 0.564 & 0.338 & 0.098 & 0.220 & 0.746 & 0.034 \\
\rowcolor{rowgray}
              & organization       & 0.308 & 0.551 & 0.371 & 0.078 & 0.214 & 0.735 & 0.051 \\
              & work of art        & 0.296 & 0.484 & 0.339 & 0.177 & 0.213 & 0.684 & 0.103 \\
\rowcolor{rowgray}
              & location           & 0.278 & 0.505 & 0.350 & 0.145 & 0.192 & 0.728 & 0.080 \\
              & person             & 0.186 & 0.353 & 0.495 & 0.152 & 0.126 & 0.797 & 0.077 \\
\midrule
\rowcolor{rowgray}
gpt-5.4       & event              & 0.542 & 0.857 & 0.076 & 0.067 & 0.396 & 0.538 & 0.066 \\
              & scientific concept & 0.433 & 0.843 & 0.116 & 0.041 & 0.291 & 0.684 & 0.025 \\
\rowcolor{rowgray}
              & work of art        & 0.423 & 0.893 & 0.079 & 0.028 & 0.277 & 0.706 & 0.017 \\
              & organization       & 0.414 & 0.876 & 0.086 & 0.038 & 0.271 & 0.696 & 0.033 \\
\rowcolor{rowgray}
              & cultural concept   & 0.383 & 0.840 & 0.127 & 0.033 & 0.248 & 0.732 & 0.020 \\
              & animal             & 0.354 & 0.874 & 0.065 & 0.061 & 0.222 & 0.740 & 0.038 \\
\rowcolor{rowgray}
              & location           & 0.330 & 0.816 & 0.092 & 0.092 & 0.207 & 0.760 & 0.033 \\
              & artifact           & 0.323 & 0.839 & 0.107 & 0.054 & 0.200 & 0.772 & 0.028 \\
\rowcolor{rowgray}
              & plant              & 0.284 & 0.921 & 0.053 & 0.026 & 0.168 & 0.821 & 0.011 \\
              & person             & 0.232 & 0.816 & 0.117 & 0.067 & 0.135 & 0.852 & 0.013 \\
\midrule
\rowcolor{rowgray}
llama-4-scout & event              & 0.172 & 0.582 & 0.208 & 0.209 & 0.101 & 0.811 & 0.088 \\
              & scientific concept & 0.109 & 0.527 & 0.237 & 0.237 & 0.061 & 0.899 & 0.040 \\
\rowcolor{rowgray}
              & work of art        & 0.095 & 0.398 & 0.293 & 0.309 & 0.054 & 0.857 & 0.089 \\
              & organization       & 0.088 & 0.530 & 0.329 & 0.141 & 0.048 & 0.923 & 0.029 \\
\rowcolor{rowgray}
              & location           & 0.085 & 0.449 & 0.339 & 0.212 & 0.047 & 0.914 & 0.039 \\
              & artifact           & 0.081 & 0.381 & 0.241 & 0.377 & 0.045 & 0.869 & 0.086 \\
\rowcolor{rowgray}
              & animal             & 0.065 & 0.710 & 0.144 & 0.146 & 0.034 & 0.946 & 0.020 \\
              & plant              & 0.063 & 0.705 & 0.205 & 0.089 & 0.033 & 0.953 & 0.014 \\
\rowcolor{rowgray}
              & cultural concept   & 0.053 & 0.469 & 0.404 & 0.127 & 0.028 & 0.949 & 0.023 \\
              & person             & 0.042 & 0.429 & 0.329 & 0.242 & 0.022 & 0.941 & 0.037 \\
\bottomrule
\end{tabular}
\end{adjustbox}
\caption{Full results for Experiment 3. Sorted by model and F1 Score in descending order.}
\label{app:full_results_exp3}
\end{table}

\begin{table}[!htbp]
\centering
\renewcommand{\arraystretch}{0.9}
\setlength{\tabcolsep}{2pt}
\scriptsize
\begin{adjustbox}{max width=\linewidth}
\begin{tabular}{l l S[table-format=1.3] S[table-format=1.3] S[table-format=1.3] S[table-format=1.3] S[table-format=1.3] S[table-format=1.3] S[table-format=1.3]}
\toprule
\rowcolor{headerblue}
\textcolor{white}{\textbf{Model}} &
\textcolor{white}{\textbf{Prompt}} &
{\textcolor{white}{\textbf{F1 (Ent.)}}} &
\multicolumn{3}{c}{\textcolor{white}{\textbf{Precision Exp.}}} &
\multicolumn{3}{c}{\textcolor{white}{\textbf{Recall Exp.}}} \\
\cmidrule(lr){4-6}\cmidrule(lr){7-9}
\rowcolor{headerblue}
& & &
{\textcolor{white}{\textbf{E}}} &
{\textcolor{white}{\textbf{N}}} &
{\textcolor{white}{\textbf{C}}} &
{\textcolor{white}{\textbf{E}}} &
{\textcolor{white}{\textbf{N}}} &
{\textcolor{white}{\textbf{C}}} \\
\midrule
\rowcolor{rowgray}
deepseek-v3.2 & GPTKB                    & 0.361 & 0.567 & 0.301 & 0.132 & 0.265 & 0.654 & 0.081 \\
              & Schema.org (no constr.)  & 0.327 & 0.578 & 0.322 & 0.100 & 0.228 & 0.721 & 0.051 \\
\rowcolor{rowgray}
              & LMCRAWL                  & 0.305 & 0.392 & 0.455 & 0.153 & 0.250 & 0.669 & 0.081 \\
              & Wikidata (no constr.)    & 0.244 & 0.495 & 0.372 & 0.133 & 0.162 & 0.792 & 0.046 \\
\rowcolor{rowgray}
              & Schema.org               & 0.216 & 0.750 & 0.164 & 0.086 & 0.126 & 0.853 & 0.021 \\
              & Wikidata                 & 0.189 & 0.715 & 0.162 & 0.123 & 0.109 & 0.866 & 0.025 \\
\midrule
\rowcolor{rowgray}
gpt-5.4       & GPTKB                    & 0.366 & 0.714 & 0.172 & 0.114 & 0.246 & 0.696 & 0.058 \\
              & Schema.org (no constr.)  & 0.338 & 0.709 & 0.185 & 0.106 & 0.222 & 0.735 & 0.043 \\
\rowcolor{rowgray}
              & Wikidata (no constr.)    & 0.310 & 0.738 & 0.165 & 0.097 & 0.196 & 0.763 & 0.041 \\
              & LMCRAWL                  & 0.296 & 0.517 & 0.337 & 0.146 & 0.207 & 0.710 & 0.083 \\
\rowcolor{rowgray}
              & Wikidata                 & 0.220 & 0.825 & 0.084 & 0.091 & 0.127 & 0.847 & 0.026 \\
              & Schema.org               & 0.205 & 0.839 & 0.088 & 0.073 & 0.117 & 0.869 & 0.014 \\
\midrule
\rowcolor{rowgray}
llama-4-scout & LMCRAWL                  & 0.198 & 0.439 & 0.361 & 0.200 & 0.128 & 0.805 & 0.067 \\
              & Schema.org (no constr.)  & 0.164 & 0.557 & 0.261 & 0.182 & 0.096 & 0.859 & 0.045 \\
\rowcolor{rowgray}
              & Wikidata (no constr.)    & 0.147 & 0.482 & 0.364 & 0.154 & 0.087 & 0.877 & 0.036 \\
              & Wikidata                 & 0.139 & 0.586 & 0.195 & 0.219 & 0.079 & 0.889 & 0.032 \\
\rowcolor{rowgray}
              & Schema.org               & 0.124 & 0.635 & 0.220 & 0.145 & 0.069 & 0.903 & 0.028 \\
              & GPTKB                    & 0.114 & 0.517 & 0.223 & 0.260 & 0.064 & 0.890 & 0.046 \\
\bottomrule
\end{tabular}
\end{adjustbox}
\caption{Full results for Experiment 4. Sorted by model and F1 Score in descending order.}
\label{app:full_results_exp4}
\end{table}

\begin{table}[!htbp]
\centering
\renewcommand{\arraystretch}{0.9}
\setlength{\tabcolsep}{2pt}
\scriptsize
\begin{adjustbox}{max width=\linewidth}
\begin{tabular}{l l S[table-format=1.3] S[table-format=1.3] S[table-format=1.3] S[table-format=1.3] S[table-format=1.3] S[table-format=1.3] S[table-format=1.3]}
\toprule
\rowcolor{headerblue}
\textcolor{white}{\textbf{Model}} &
\textcolor{white}{\textbf{Triple Range (pop / lt)}} &
{\textcolor{white}{\textbf{F1 (Ent.)}}} &
\multicolumn{3}{c}{\textcolor{white}{\textbf{Precision Exp.}}} &
\multicolumn{3}{c}{\textcolor{white}{\textbf{Recall Exp.}}} \\
\cmidrule(lr){4-6}\cmidrule(lr){7-9}
\rowcolor{headerblue}
& & &
{\textcolor{white}{\textbf{E}}} &
{\textcolor{white}{\textbf{N}}} &
{\textcolor{white}{\textbf{C}}} &
{\textcolor{white}{\textbf{E}}} &
{\textcolor{white}{\textbf{N}}} &
{\textcolor{white}{\textbf{C}}} \\
\midrule
\rowcolor{rowgray}
deepseek-v3.2 & 33--66 / 3--7     & 0.371 & 0.550 & 0.310 & 0.140 & 0.280 & 0.626 & 0.094 \\
              & 50--100 / 5--10   & 0.358 & 0.544 & 0.303 & 0.153 & 0.267 & 0.644 & 0.089 \\
\rowcolor{rowgray}
              & 100--200 / 10--20 & 0.351 & 0.494 & 0.369 & 0.137 & 0.272 & 0.650 & 0.078 \\
              & 25--50 / 3--5     & 0.349 & 0.546 & 0.291 & 0.163 & 0.257 & 0.652 & 0.091 \\
\rowcolor{rowgray}
              & 75--150 / 8--15   & 0.348 & 0.503 & 0.370 & 0.127 & 0.266 & 0.649 & 0.085 \\
              & 10--20 / 1--2     & 0.315 & 0.638 & 0.209 & 0.153 & 0.209 & 0.722 & 0.069 \\
\rowcolor{rowgray}
              & 5--10 / 1         & 0.305 & 0.686 & 0.160 & 0.154 & 0.196 & 0.754 & 0.050 \\
\midrule
\rowcolor{rowgray}
gpt-5.4       & 33--66 / 3--7     & 0.402 & 0.720 & 0.193 & 0.087 & 0.279 & 0.667 & 0.054 \\
              & 100--200 / 10--20 & 0.396 & 0.715 & 0.177 & 0.108 & 0.274 & 0.682 & 0.043 \\
\rowcolor{rowgray}
              & 75--150 / 8--15   & 0.387 & 0.715 & 0.172 & 0.113 & 0.265 & 0.684 & 0.051 \\
              & 50--100 / 5--10   & 0.380 & 0.710 & 0.176 & 0.114 & 0.259 & 0.697 & 0.044 \\
\rowcolor{rowgray}
              & 25--50 / 3--5     & 0.367 & 0.734 & 0.143 & 0.123 & 0.245 & 0.713 & 0.042 \\
              & 5--10 / 1         & 0.327 & 0.777 & 0.119 & 0.104 & 0.207 & 0.740 & 0.053 \\
\rowcolor{rowgray}
              & 10--20 / 1--2     & 0.327 & 0.801 & 0.119 & 0.080 & 0.205 & 0.752 & 0.043 \\
\midrule
\rowcolor{rowgray}
llama-4-scout & 50--100 / 5--10   & 0.130 & 0.525 & 0.235 & 0.240 & 0.074 & 0.877 & 0.047 \\
              & 33--66 / 3--7     & 0.127 & 0.543 & 0.220 & 0.237 & 0.072 & 0.885 & 0.043 \\
\rowcolor{rowgray}
              & 75--150 / 8--15   & 0.121 & 0.484 & 0.239 & 0.277 & 0.069 & 0.875 & 0.056 \\
              & 5--10 / 1         & 0.114 & 0.601 & 0.166 & 0.233 & 0.063 & 0.909 & 0.028 \\
\rowcolor{rowgray}
              & 100--200 / 10--20 & 0.103 & 0.522 & 0.248 & 0.230 & 0.057 & 0.895 & 0.047 \\
              & 25--50 / 3--5     & 0.100 & 0.531 & 0.230 & 0.239 & 0.055 & 0.898 & 0.046 \\
\rowcolor{rowgray}
              & 10--20 / 1--2     & 0.098 & 0.555 & 0.197 & 0.248 & 0.054 & 0.908 & 0.038 \\
\bottomrule
\end{tabular}
\end{adjustbox}
\caption{Full results for Experiment 5. Sorted by model and F1 Score in descending order.}
\label{app:full_results_exp5}
\end{table}

\begin{table*}[!tbp]
\centering
\scriptsize
\renewcommand{\arraystretch}{0.9}
\setlength{\tabcolsep}{3pt}
\begin{adjustbox}{max width=\textwidth}
\begin{tabular}{lrrrrrrr}
\toprule
\textbf{Model} & \textbf{Avg} & \textbf{Total} & \textbf{Uniq.} & \textbf{Uniq.} & \textbf{Uniq.} & \textbf{Max} & \textbf{Min} \\
 & \textbf{Tri./Subj.} & \textbf{Tri.} & \textbf{Subj.} & \textbf{Pred.} & \textbf{Obj.} & \textbf{Tri./Subj.} & \textbf{Tri./Subj.} \\
\midrule
Qwen 3.5 27B             &  60.05 & 30023 & 420 & 9718 & 18221 & 2101 & 1 \\
Gemma 3 12B              &  52.40 & 26198 & 1142 & 4659 & 20142 & 388 & 1 \\
Gemini 3 Flash           &  44.75 & 22375 & 501 & 4153 & 17920 & 122 & 1 \\
GPT OSS 120B             &  41.50 & 20749 & 567 & 7585 & 16745 & 197 & 1 \\
DeepSeek V3.2            &  37.04 & 18519 & 492 & 5883 & 13896 & 695 & 1 \\
Gemma 3 27B              &  32.67 & 16335 & 651 & 3089 & 13412 & 97 & 1 \\
Kimi K2.5                &  31.03 & 15513 & 462 & 3701 & 11552 & 238 & 4 \\
MiniMax M2.5             &  22.06 & 11031 & 423 & 2954 & 9001 & 1819 & 1 \\
GPT-5.4                  &  20.08 & 10038 & 450 & 2883 & 7505 & 97 & 1 \\
Gemini 3.1 Flash Lite    &  18.97 & 9487 & 505 & 1904 & 7087 & 56 & 1 \\
Mistral Large 2512       &  18.71 & 9354 & 505 & 1818 & 7485 & 153 & 1 \\
Claude Opus 4.6          &  18.70 & 9348 & 585 & 2500 & 7213 & 72 & 1 \\
Claude Sonnet 4.6        &  16.41 & 8205 & 496 & 2238 & 6526 & 76 & 1 \\
GPT-5 Mini               &  15.78 & 7888 & 316 & 3219 & 6764 & 98 & 1 \\
Gemma 3 4B               &  15.56 & 7779 & 483 & 1980 & 6383 & 139 & 1 \\
Gemini 3.1 Pro           &  12.93 & 6464 & 471 & 921 & 4953 & 65 & 4 \\
Claude Haiku 4.5         &   8.68 & 4340 & 378 & 1300 & 3451 & 51 & 1 \\
Grok 4.3                 &   7.44 & 3719 & 277 & 999 & 2997 & 52 & 3 \\
Llama 4 Scout 17B        &   6.41 & 3205 & 341 & 764 & 2400 & 211 & 1 \\
GPT-5 Nano               &   4.17 & 2083 & 366 & 697 & 1662 & 49 & 1 \\
\bottomrule
\end{tabular}
\end{adjustbox}
\caption{Statistics of elicited triples in Experiment 1, averaged over the 5 runs. Sorted by average triples per subject (desc.). Note: avg tri./subj is computed on the whole set of entities for this experiment (500), but some models did not produce triples for some entities or added entities not queried for.}
\label{app:elicit_stats_exp1}
\end{table*}

\begin{table*}[!tbp]
\centering
\scriptsize
\renewcommand{\arraystretch}{0.9}
\setlength{\tabcolsep}{3pt}
\begin{adjustbox}{max width=\textwidth}
\begin{tabular}{llrrrrrrr}
\toprule
\textbf{Model} & \textbf{Eff.} & \textbf{Avg} & \textbf{Total} & \textbf{Uniq.} & \textbf{Uniq.} & \textbf{Uniq.} & \textbf{Max} & \textbf{Min} \\
 & & \textbf{Tri./Subj.} & \textbf{Tri.} & \textbf{Subj.} & \textbf{Pred.} & \textbf{Obj.} & \textbf{Tri./Subj.} & \textbf{Tri./Subj.} \\
\midrule
Claude Opus 4.6    & high                      &  23.32 & 11658 & 752 & 3196 & 8877 & 95 & 1 \\
GPT-5.4            & high                      &  22.90 & 11449 & 454 & 3527 & 8664 & 109 & 2 \\
GPT-5.4            & low                       &  21.48 & 10738 & 456 & 3075 & 7965 & 99 & 1 \\
GPT-5.4            & med                       &  20.93 & 10467 & 452 & 3055 & 7782 & 94 & 1 \\
Claude Opus 4.6    & med                       &  18.63 & 9315 & 575 & 2477 & 7221 & 69 & 1 \\
Gemini 3.1 Pro     & high                      &  14.56 & 7282 & 475 & 1040 & 5624 & 53 & 5 \\
Gemini 3.1 Pro     & med                       &  12.98 & 6488 & 476 & 930 & 4976 & 68 & 4 \\
Claude Opus 4.6    & low                       &  11.00 & 5502 & 441 & 1464 & 4294 & 64 & 1 \\
Gemini 3.1 Pro     & low                       &  10.53 & 5267 & 483 & 721 & 3902 & 33 & 3 \\
\bottomrule
\end{tabular}
\end{adjustbox}
\caption{Statistics of elicited triples in Experiment 2. Sorted by average triples per subject (desc.). Note: avg tri./subj is computed on the whole set of entities for this experiment (500), but some models did not produce triples for some entities or added entities not queried for.}
\label{app:elicit_stats_exp2}
\end{table*}

\begin{table*}[!tbp]
\centering
\scriptsize
\renewcommand{\arraystretch}{0.9}
\setlength{\tabcolsep}{3pt}
\begin{adjustbox}{max width=\textwidth}
\begin{tabular}{lrrrrrrr}
\toprule
\textbf{Model} & \textbf{Avg} & \textbf{Total} & \textbf{Uniq.} & \textbf{Uniq.} & \textbf{Uniq.} & \textbf{Max} & \textbf{Min} \\
 & \textbf{Tri./Subj.} & \textbf{Tri.} & \textbf{Subj.} & \textbf{Pred.} & \textbf{Obj.} & \textbf{Tri./Subj.} & \textbf{Tri./Subj.} \\
\midrule
DeepSeek V3.2                                  &  37.39 & 37393 & 1043 & 8910 & 27366 & 928 & 1 \\
GPT-5.4                                        &  19.83 & 19832 & 900 & 4864 & 13588 & 99 & 2 \\
Llama 4 Scout 17B                              &   6.32 & 6322 & 721 & 1231 & 4453 & 377 & 1 \\
\bottomrule
\end{tabular}
\end{adjustbox}
\caption{Statistics of elicited triples in Experiment 3. Sorted by average triples per subject (desc.). Note: avg tri./subj is computed on the whole set of entities for this experiment (1000), but some models did not produce triples for some entities or added entities not queried for.}
\label{app:elicit_stats_exp3}
\end{table*}

\begin{table*}[!tbp]
\centering
\scriptsize
\renewcommand{\arraystretch}{0.9}
\setlength{\tabcolsep}{3pt}
\begin{adjustbox}{max width=\textwidth}
\begin{tabular}{llrrrrrrr}
\toprule
\textbf{Model} & \textbf{Prompt} & \textbf{Avg} & \textbf{Total} & \textbf{Uniq.} & \textbf{Uniq.} & \textbf{Uniq.} & \textbf{Max} & \textbf{Min} \\
 & & \textbf{Tri./Subj.} & \textbf{Tri.} & \textbf{Subj.} & \textbf{Pred.} & \textbf{Obj.} & \textbf{Tri./Subj.} & \textbf{Tri./Subj.} \\
\midrule
DeepSeek V3.2      & Wikidata (free)           & 178.19 & 89094 & 435 & 8951 & 77126 & 4095 & 1 \\
DeepSeek V3.2      & LMCRAWL                   & 172.31 & 86154 & 446 & 7417 & 66948 & 9247 & 1 \\
Llama 4 Scout 17B  & LMCRAWL                   &  74.12 & 37059 & 493 & 2634 & 24874 & 6419 & 1 \\
DeepSeek V3.2      & schema.org (free)         &  64.21 & 32107 & 453 & 1039 & 28848 & 3141 & 1 \\
DeepSeek V3.2      & GPTKB                     &  38.49 & 19243 & 519 & 4883 & 14991 & 255 & 1 \\
GPT-5.4            & LMCRAWL                   &  31.43 & 15717 & 500 & 1521 & 12023 & 133 & 1 \\
GPT-5.4            & GPTKB                     &  25.04 & 12519 & 501 & 3347 & 9342 & 100 & 1 \\
GPT-5.4            & schema.org (free)         &  22.11 & 11054 & 457 & 432 & 8733 & 81 & 1 \\
DeepSeek V3.2      & schema.org (constr.)      &  15.00 & 7498 & 408 & 1286 & 5111 & 854 & 1 \\
GPT-5.4            & Wikidata (free)           &  14.38 & 7189 & 470 & 572 & 5289 & 61 & 1 \\
Llama 4 Scout 17B  & schema.org (free)         &  13.23 & 6615 & 396 & 1562 & 4442 & 387 & 1 \\
Llama 4 Scout 17B  & Wikidata (free)           &  12.99 & 6496 & 400 & 1713 & 4532 & 248 & 1 \\
DeepSeek V3.2      & Wikidata (constr.)        &   9.65 & 4827 & 355 & 376 & 3597 & 257 & 1 \\
Llama 4 Scout 17B  & Wikidata (constr.)        &   8.57 & 4284 & 407 & 248 & 2281 & 274 & 1 \\
Llama 4 Scout 17B  & schema.org (constr.)      &   7.85 & 3924 & 408 & 114 & 2514 & 246 & 1 \\
GPT-5.4            & schema.org (constr.)      &   7.65 & 3823 & 465 & 41 & 2847 & 27 & 1 \\
GPT-5.4            & Wikidata (constr.)        &   6.91 & 3457 & 463 & 64 & 2271 & 46 & 1 \\
Llama 4 Scout 17B  & GPTKB                     &   6.32 & 3162 & 334 & 770 & 2346 & 224 & 1 \\
\bottomrule
\end{tabular}
\end{adjustbox}
\caption{Statistics of elicited triples in Experiment 4. Sorted by average triples per subject (desc.). Note: avg tri./subj is computed on the whole set of entities for this experiment (500), but some models did not produce triples for some entities or added entities not queried for.}
\label{app:elicit_stats_exp4}
\end{table*}

\begin{table*}[!tbp]
\centering
\scriptsize
\renewcommand{\arraystretch}{0.9}
\setlength{\tabcolsep}{3pt}
\begin{adjustbox}{max width=\textwidth}
\begin{tabular}{llrrrrrrr}
\toprule
\textbf{Model} & \textbf{Range} & \textbf{Avg} & \textbf{Total} & \textbf{Uniq.} & \textbf{Uniq.} & \textbf{Uniq.} & \textbf{Max} & \textbf{Min} \\
 & & \textbf{Tri./Subj.} & \textbf{Tri.} & \textbf{Subj.} & \textbf{Pred.} & \textbf{Obj.} & \textbf{Tri./Subj.} & \textbf{Tri./Subj.} \\
\midrule
DeepSeek V3.2      & pop 100--200 / lt 10--20  &  56.13 & 28063 & 469 & 8261 & 21852 & 504 & 1 \\
DeepSeek V3.2      & pop 75--150 / lt 8--15    &  51.34 & 25671 & 517 & 6480 & 20418 & 729 & 1 \\
DeepSeek V3.2      & pop 33--66 / lt 3--7      &  38.45 & 19227 & 544 & 4807 & 15084 & 144 & 1 \\
DeepSeek V3.2      & pop 50--100 / lt 5--10    &  36.00 & 18001 & 504 & 4491 & 14339 & 293 & 1 \\
GPT-5.4            & pop 75--150 / lt 8--15    &  32.64 & 16321 & 501 & 4506 & 12111 & 122 & 1 \\
DeepSeek V3.2      & pop 25--50 / lt 3--5      &  28.43 & 14217 & 507 & 3503 & 11233 & 125 & 1 \\
GPT-5.4            & pop 100--200 / lt 10--20  &  26.01 & 13005 & 493 & 3437 & 9668 & 170 & 1 \\
GPT-5.4            & pop 50--100 / lt 5--10    &  24.05 & 12024 & 521 & 3088 & 9009 & 97 & 1 \\
GPT-5.4            & pop 33--66 / lt 3--7      &  23.78 & 11890 & 497 & 3116 & 8866 & 68 & 1 \\
GPT-5.4            & pop 25--50 / lt 3--5      &  19.47 & 9737 & 495 & 2451 & 7340 & 58 & 1 \\
DeepSeek V3.2      & pop 10--20 / lt 1--2      &  16.42 & 8212 & 489 & 1906 & 6527 & 38 & 1 \\
GPT-5.4            & pop 10--20 / lt 1--2      &  13.43 & 6715 & 497 & 1548 & 4993 & 47 & 1 \\
GPT-5.4            & pop 5--10 / lt 1          &  11.44 & 5720 & 495 & 1249 & 4272 & 44 & 1 \\
DeepSeek V3.2      & pop 5--10 / lt 1          &  11.30 & 5652 & 484 & 1271 & 4512 & 27 & 1 \\
Llama 4 Scout 17B  & pop 75--150 / lt 8--15    &   8.19 & 4096 & 386 & 1058 & 3025 & 237 & 1 \\
Llama 4 Scout 17B  & pop 100--200 / lt 10--20  &   7.25 & 3625 & 363 & 868 & 2666 & 232 & 1 \\
Llama 4 Scout 17B  & pop 50--100 / lt 5--10    &   6.74 & 3371 & 348 & 827 & 2511 & 232 & 1 \\
Llama 4 Scout 17B  & pop 33--66 / lt 3--7      &   6.05 & 3023 & 338 & 710 & 2217 & 257 & 1 \\
Llama 4 Scout 17B  & pop 25--50 / lt 3--5      &   5.79 & 2893 & 298 & 748 & 2135 & 235 & 1 \\
Llama 4 Scout 17B  & pop 10--20 / lt 1--2      &   4.94 & 2470 & 297 & 574 & 1833 & 221 & 1 \\
Llama 4 Scout 17B  & pop 5--10 / lt 1          &   4.19 & 2097 & 350 & 460 & 1561 & 173 & 1 \\
\bottomrule
\end{tabular}
\end{adjustbox}
\caption{Statistics of elicited triples in Experiment 5. Sorted by average triples per subject (desc.). Note: avg tri./subj is computed on the whole set of entities for this experiment (500), but some models did not produce triples for some entities or added entities not queried for.}
\label{app:elicit_stats_exp5}
\end{table*}
\end{document}